\documentclass[sigconf,9pt]{acmart}

\makeatletter
\def\@ACM@checkaffil{
    \if@ACM@instpresent\else
    \ClassWarningNoLine{\@classname}{No institution present for an affiliation}%
    \fi
    \if@ACM@citypresent\else
    \ClassWarningNoLine{\@classname}{No city present for an affiliation}%
    \fi
    \if@ACM@countrypresent\else
        \ClassWarningNoLine{\@classname}{No country present for an affiliation}%
    \fi
}
\makeatother

\AtBeginDocument{%
  \providecommand\BibTeX{{%
    \normalfont B\kern-0.5em{\scshape i\kern-0.25em b}\kern-0.8em\TeX}}}

\setcopyright{acmcopyright}
\copyrightyear{2025}
\acmYear{2025}
\setcopyright{rightsretained}
\acmConference[ACM SenSys '25]{The 23th ACM Conference on Embedded Networked Sensor Systems}{May 6-9, 2025}{Irvine, USA}
\acmBooktitle{The 23th ACM Conference on Embedded Networked Sensor Systems (SenSys '25), May 6-9, 2025, Irvine, USA}
\acmDOI{xx.xxxx/xxxxxxx.xxxxxxx}
\acmISBN{xxx-x-xxxx-xxxx-x/xx/xx}

\usepackage[ruled,linesnumbered]{algorithm2e}
\usepackage{subfigure}
\usepackage{multicol}
\usepackage{algorithmic}
\usepackage{caption}
\usepackage{bbding}
\usepackage{pifont}
\usepackage{graphicx} 
\usepackage{url}
\usepackage{gensymb}

\ifodd 1
\newcommand{\com}[1]{\textbf{\color{red}(COMMENT: #1)}} 
\else

\newcommand{\com}[1]{}
\fi

\newcommand{\revise}[1]{{\color{black}{#1}}}

\def\fig{Fig.}

\def\eg{e.g.}
\def\ie{i.e.}
\def\aka{a.k.a.}
\def\eqn{Eqn.}

\def\alg{Algorithm}

\begin{document}

\newcommand{\mycustomsize}{\fontsize{21}{\baselineskip}\selectfont}

\title[Ultra-High-Frequency Harmony: \\mmWave Radar and Event Camera Orchestrate Accurate Drone Landing]
{Ultra-High-Frequency Harmony: mmWave Radar and \\ Event Camera Orchestrate Accurate Drone Landing }

\author{Haoyang Wang$^{1}$, Jingao Xu$^{2}$, Xinyu Luo$^{1}$, Xuecheng Chen$^{1}$, Ting Zhang$^{1}$, \\ Ruiyang Duan$^3$, Yunhao Liu$^{4}$, Xinlei Chen$^{1, 5, 6}$\textsuperscript{\Envelope}}

\renewcommand{\authors}{Haoyang Wang, Jingao Xu, Xinyu Luo, Xuecheng Chen, Ting Zhang, Ruiyang Duan, Yunhao Liu, Xinlei Chen}

\affiliation{%
    \institution{$^1$ Shenzhen International Graduate School, Tsinghua University, China; $^2$ Carnegie Mellon University; \\$^3$ Meituan Academy of Robotics Shenzhen, China; $^4$ School of Software, Tsinghua University, China; \\$^5$ Pengcheng Laboratory, Shenzhen, China; $^6$ RISC-V International Open Source Laboratory, Shenzhen, China}
    \country{}
    \city{}
}


\affiliation{%
  \institution{Email: \{haoyang-22, luo-xy23, chenxc21, yunhao\}@mails.tsinghua.edu.cn, \{xujingao13, zhangt2112\}@gmail.com, \\ duanruiyang@meituan.com, chen.xinlei@sz.tsinghua.edu.cn}
}

\renewcommand{\shortauthors}{Haoyang Wang, et al.}

\begin{abstract}

For precise, efficient, and safe drone landings, ground platforms should real-time, accurately locate descending drones and guide them to designated spots.
While mmWave sensing combined with cameras improves localization accuracy, the lower sampling frequency of traditional frame cameras compared to mmWave radar creates bottlenecks in system throughput. 
In this work, we replace the traditional frame camera with event camera, a novel sensor that harmonizes in sampling frequency with mmWave radar within the ground platform setup, and introduce mmE-Loc, a high-precision, low-latency ground localization system designed for drone landings.
To fully leverage the \textit{temporal consistency} and \textit{spatial complementarity} between these modalities, we propose two innovative modules, \textit{consistency-instructed collaborative tracking} and \textit{graph-informed adaptive joint optimization}, for accurate drone measurement extraction and efficient sensor fusion.
Extensive real-world experiments in landing scenarios from a leading drone delivery company demonstrate that mmE-Loc outperforms state-of-the-art methods in both localization accuracy and latency. 
\end{abstract}


%

\maketitle 

\renewcommand{\thefootnote}{}
\footnotetext{\Envelope\ Corresponding author.}
\footnotetext{Project homepage: \href{https://mmE-Loc.github.io/}{\color{blue}{https://mmE-Loc.github.io/}}}

\vspace{-0.8cm}
\section{Introduction}

\begin{figure}[t]
    \setlength{\abovecaptionskip}{0.25cm} 
    \setlength{\belowcaptionskip}{-0.3cm}
    \setlength{\subfigcapskip}{-0.5cm}
    \centering
        \includegraphics[width=0.92\columnwidth]{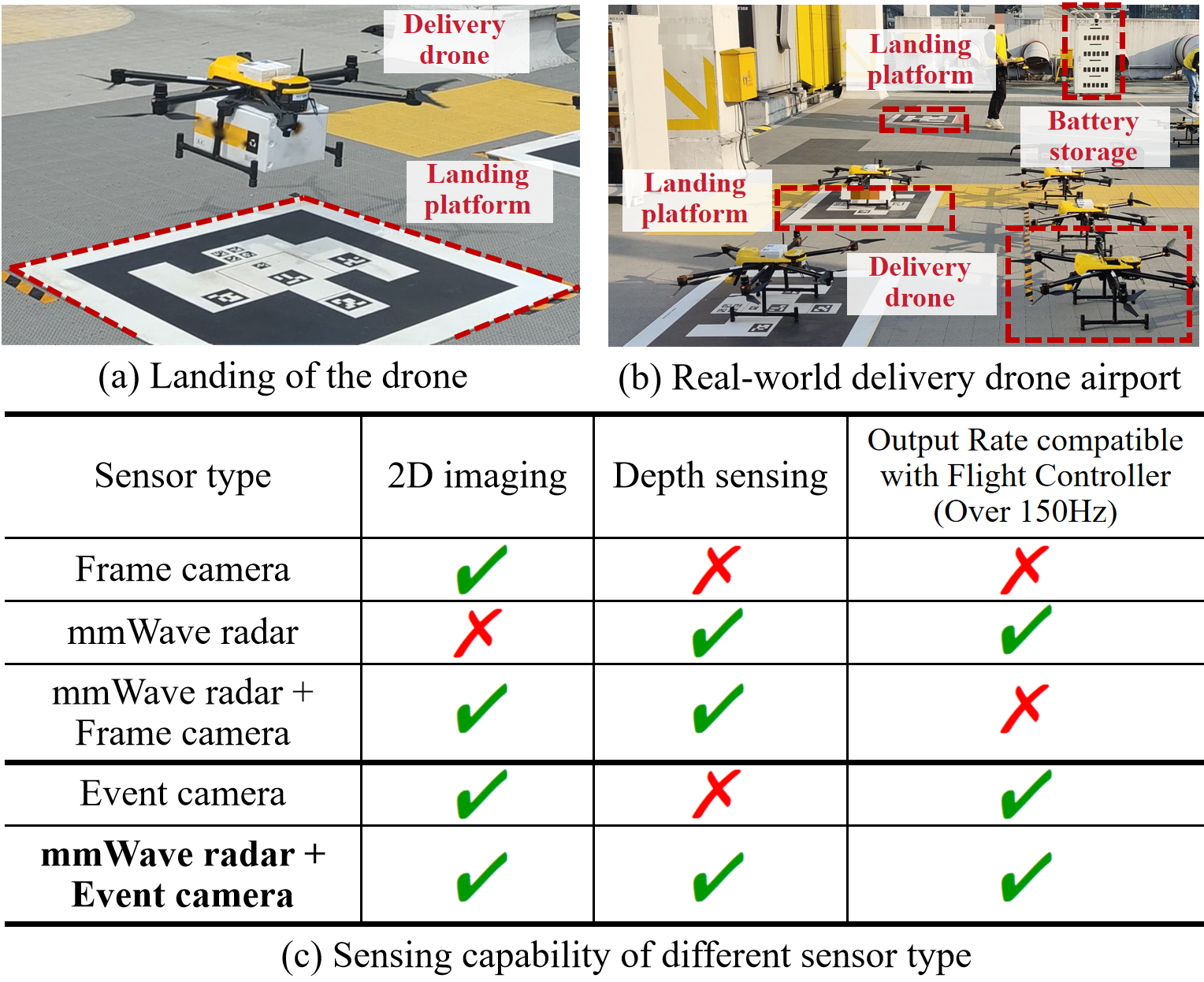}
        \vspace{-0.2cm}
    \caption{Snapshot of drone landing phase, airport, and sensors performance. \textnormal{(a) A delivery drone lands on the platform. (b) The real-world drone airport is equipped with multiple drones for package delivery. 
    (c) Integrating mmWave radar with event camera combines reliable depth sensing and 2D imaging at ultra-high sampling frequencies, enabling high spatial-temporal resolution and depth sensing, while ensuring compatibility with flight controllers.}}
    \label{intro}
\end{figure}

Projected to soar to a \$1 trillion market by 2040 \cite{low_altitude_eco}, the drone-driven low-altitude economy is transforming sectors with revolutionary applications such as on-demand delivery \cite{wang2022micnest, chen2024ddl, chen2022deliversense}, meticulous industrial inspections \cite{xu2022swarmmap, li2024quest, xu2019vehicle, liu2024mobiair}, and rapid relief-and-rescue \cite{zhang2023rf, chi2022wi, chen2024soscheduler}. 
Of paramount importance within this burgeoning sector is the \textit{landing phase}, where ground platforms locate drones descending from below 10 meters and guide them to accurately land at designated spots (\fig\ref{intro}a) \cite{he2023acoustic, sun2022aim}.
Situated near populated and commercial zones, these operations emphasize safety and reliability: our research with a leading drone delivery company reveals that a landing bias of just 10$cm$ will result in drones damaging delivery targets or missing their charging ports \cite{gonzalez2021visual}. 
Such inaccuracies disrupt the operational efficiency of this swiftly growing economic sector, with potentially severe consequences.

Widely adopted and straightforward approaches involve installing cameras at the center or edges of drone landing pads and employing computer vision algorithms for drone localization \cite{li2018real,xia2023anemoi,zhang2019eye}.
However, traditional frame cameras' Achilles heel is capturing only 2D images without depth information, leading to scale uncertainty that limits the 3D localization accuracy \cite{zhang2022mobidepth,xie2023mozart,10.1145/3517260}.
To address this shortcoming, current practices have incorporated mmWave sensing to provide the lacked depth information for better localization accuracy and reliability in various conditions \cite{deng2022geryon, lu2020smokerobustindoormapping,zhang2023mmhawkeye, sie2023batmobility, iizuka2023millisign, lu2020see, lu2020milliego}.

Albeit inspiring, our benchmark study with a world-leading drone delivery company in landing scenarios (\fig \ref{intro}b) reveals another critical drawback (\fig \ref{intro}c): the exposure times of frame cameras (>20$ms$) prevent their sampling rates from matching the high frequency of mmWave radars (\eg, 200Hz). 
This limitation creates system efficiency and throughput bottlenecks, restricting drone location updates to below 50Hz.
In contrast, drone flight controllers typically require location input rates over 150Hz to precisely adjust the drone's flight attitude for safe landing \cite{8412592, 10.1145/3570361.3592532}.
The inefficiency originates from the inherent physical limitations of conventional frame cameras and cannot be easily solved by software solutions.

\noindent \textbf{Upgrade frame camera to event camera.}
Event cameras are bio-inspired sensors that report pixel-wise intensity changes with $ms$-level resolution \cite{gallego2020event, ruan2024distill}, capturing high-speed motions without blurring \cite{he2024microsaccade}, ideal for fast-tracking tasks \cite{xu2023taming, luo2024eventtracker}.
Event cameras offer $ms$-level sampling latency, which harmonizes exceptionally with the high sampling frequencies of mmWave radar.
Their 2D imaging capability also complements radars' limited spatial resolution, similar to how traditional frame cameras operate.
Such \textit{temporal-consistency} and \textit{spatial-complementarity} across both modalities inspire us to upgrade frame cameras with event cameras to pair with radar for accurate and fast drone localization.

\noindent \textbf{Our work.}
Following the above insight, we present \textbf{mmE-Loc}, the first active, high-precision, and low-latency landing drone ground localization system that enhances mmWave radar functionality with event cameras. 
mmE-Loc works in scenarios where urban canyon environments degrade the accuracy of GPS or RTK systems as altitude decreases, rendering them nearly ineffective for the landing phase. 
With mmE-Loc, drones achieve reliable localization even under challenging conditions (\eg, weak illumination), ensuring stable and efficient landing.

\begin{figure}[t]
    \setlength{\belowcaptionskip}{-0.3cm}
    \setlength{\subfigcapskip}{-0.25cm}
    \centering
        \includegraphics[width=0.9\columnwidth]{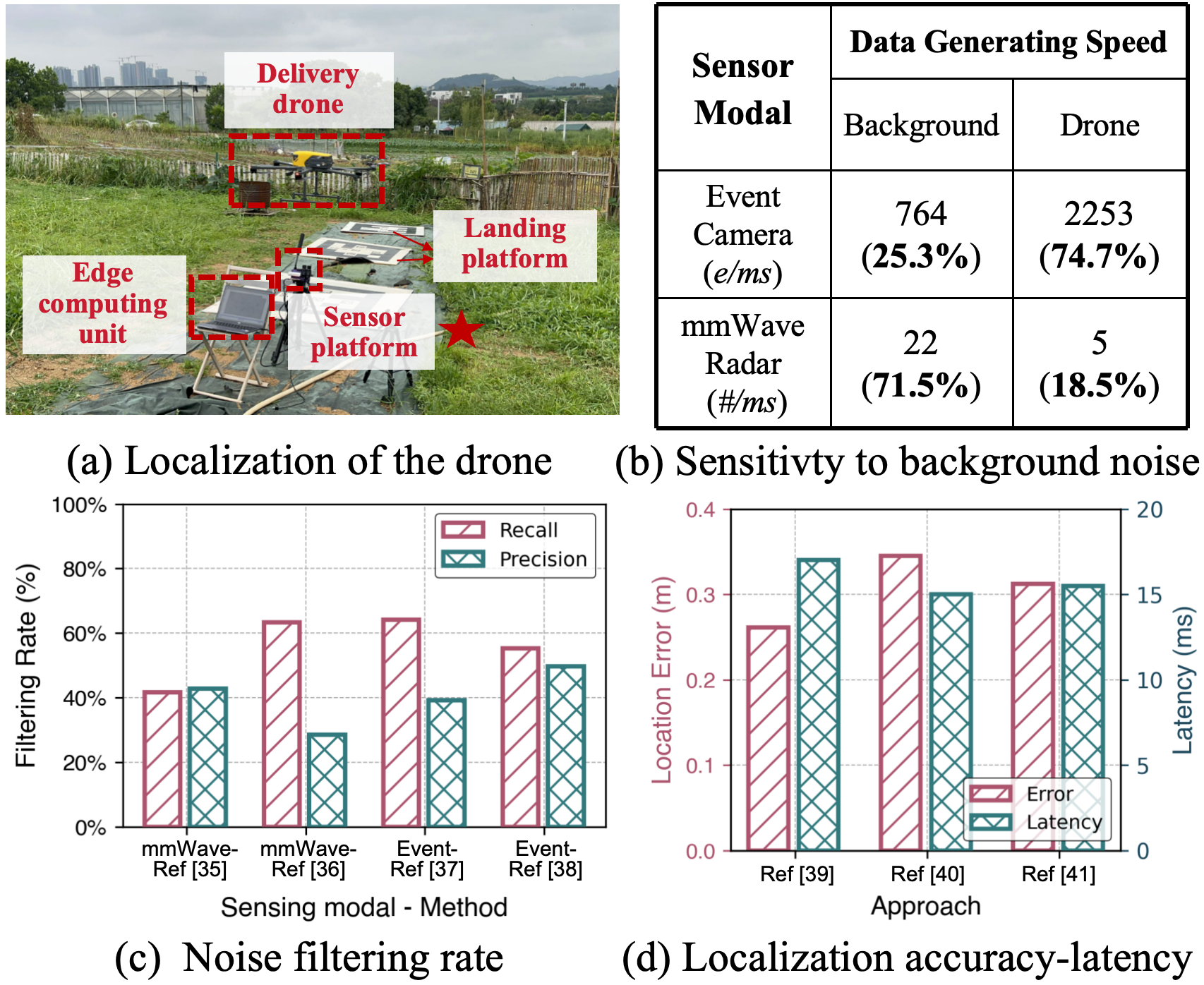}
        \vspace{-0.4cm}
    \caption{
    Benchmark study on drone localization.
    \textnormal{(a) Benchmark study at a real-world drone delivery airport; (b) Both sensors are sensitive to environmental variations; (c) Existing algorithms suffer from low noise filtering rates; (d) Existing algorithms experience cumulative drift errors and delays. }}
    \label{relatedwork}
    \vspace{-0.5cm}
\end{figure}

However, our benchmark study at a real-world drone delivery airport (\fig \ref{relatedwork}a) highlights several challenges that have been solved in making \textbf{mmE-Loc} a viable system outdoors:
$(i)$ \textit{How to accurately extract drone-related measurements} given the immense noisy output of event cameras and mmWave radars, which also lack inherent drone semantic information and differ greatly in dimension and pattern?
Both modalities are sensitive to environmental variations (\eg, changes in lighting conditions), as shown in \fig \ref{relatedwork}b.
Existing algorithms \cite{cao2024virteach, liu2024pmtrack, wang2021asynchronous, alzugaray2018asynchronous} are typically designed for single-modality, resulting in low noise filtering rates (recall and precision < 65\% in \fig \ref{relatedwork}c).
$(ii)$ \textit{How to efficiently fuse event camera and mmWave readings} that are heterogeneous in measurement precision, scale, and density? 
Existing EKF (extended Kalman filter) or PF (particle filter) based approaches \cite{falanga2020dynamic,zhao20213d, mitrokhin2018event}, suffer from cumulative drift errors, making them insufficient for precise localization (\fig \ref{relatedwork}d).
$(iii)$ \textit{How to optimize the efficiency of the fusion algorithm} to achieve high-frequency drone ground localization, given the limited computational resources on landing platforms?
Existing methods experience significant processing delays, rendering them unsuitable for low-latency localization tasks (\fig \ref{relatedwork}d) \cite{zhao20213d, falanga2020dynamic, mitrokhin2018event}.

To solve the above challenges, the design and implementation of mmE-Loc excel in the three aspects of drone ground localization:

\noindent $\bullet$ \textit{On system architecture front.}
Upgrading frame camera to event camera with $ms$-level latency to pair the mmWave radar, mmE-Loc improves drone ground localization at data source.
The system architecture tightly integrates both modalities, from early-stage noise filtering and drone detection to later-stage fusion and optimization, fully leveraging the unique advantages of both sensors (§\ref{3.2}).\\
\noindent $\bullet$ \textit{On system algorithm front.}
We introduce a Consistency-Instructed Collaborative Tracking (\textit{CCT}) algorithm, which leverages the drone's periodic micro-motion and cross-modal \textit{temporal-consistency} to filter environment-triggered noise, achieving accurate drone detection (§\ref{4.1}). 
We then present a Graph-Informed Adaptive Joint Optimization (\textit{GAJO}) algorithm, which fuses \textit{spatial-complementarity} with a novel factor graph to boost drone ground localization, resulting in a trajectory with minimal bias and low cumulative drift(§\ref{4.2}). \\
\noindent $\bullet$ \textit{On system implementation front.}
We further analyze the sources of latency and propose an Adaptive Optimization method to improve the efficiency of the \textit{GAJO} algorithm.
This approach allows \textit{GAJO} to dynamically optimize a set of locations, maintaining accuracy while reducing latency (§\ref{I}).

\begin{figure*}[t]
    \setlength{\abovecaptionskip}{0.2cm} 
    \setlength{\belowcaptionskip}{-0.5cm}
    \setlength{\subfigcapskip}{-0.25cm}
    \centering
        \includegraphics[width=1.9\columnwidth]{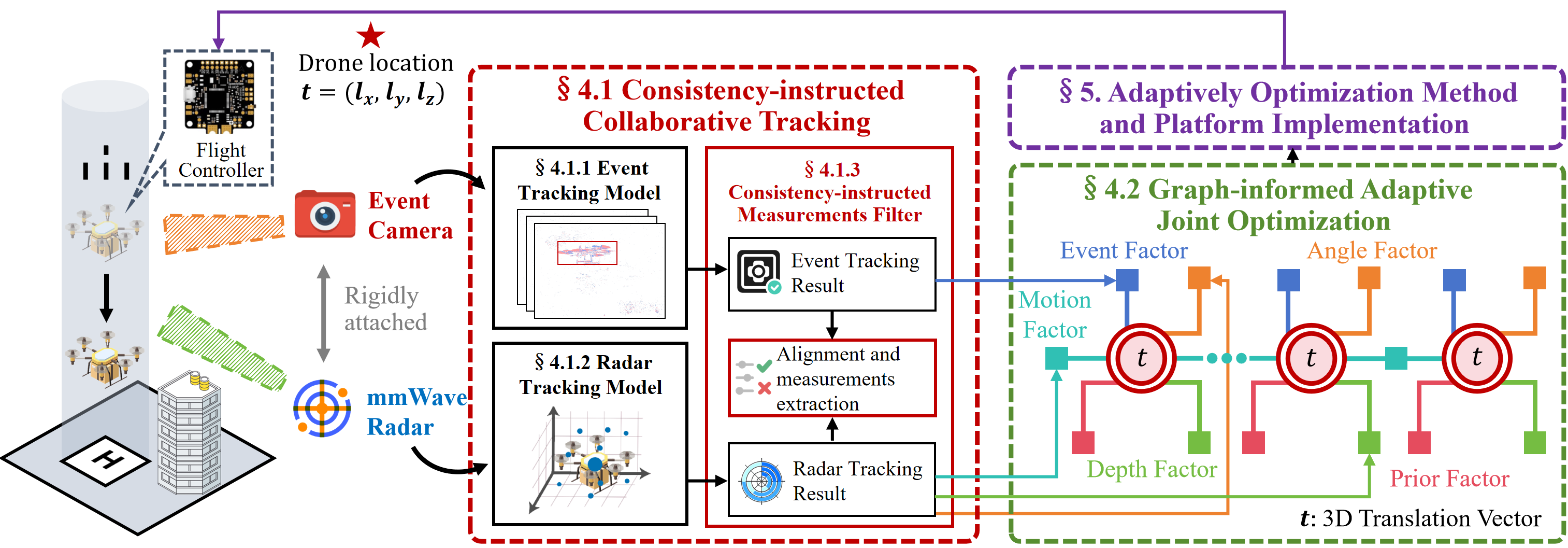}
        \vspace{-0.1cm}
    \caption{System architecture of mmE-Loc.}
    \label{overview}
\end{figure*}

We fully implement mmE-Loc using a COTS event camera and mmWave radar. 
Over 30+ hours of indoor and outdoor experiments under various drone flight conditions assess its localization accuracy and end-to-end latency performance against four SOTA methods.
mmE-Loc achieves an average localization accuracy of 0.083$m$ and latency of 5.15$ms$, surpassing baselines by >48\% and >62\%, respectively, and showing minimal sensitivity to drone type and environment.
We also deploy mmE-Loc at a real-world drone delivery airport (\fig \ref{relatedwork}a) for 10 hours, demonstrating its practicality for commercial-level drone landing requirements. 

In summary, this paper makes the following contributions.\\
\noindent $(1)$ We explore a fresh sensor configuration, event camera plus mmWave radar, that embraces and harmonizes ultra-high sampling frequencies and propose mmE-Loc, a ground localization system for drone landings that delivers $cm$ accuracy and $ms$ latency.\\
\noindent $(2)$ We present $CCT$, which leverages \textit{temporal consistency} and the drone's periodic micro-motions for precise drone detection; and $GAJO$, which employs \textit{spatial complementarity} with a novel factor graph to enhance drone localization.\\
\noindent $(3)$ We implement and extensively evaluate mmE-Loc by comparing it with four SOTA methods, showing its effectiveness. We also deploy mmE-Loc in a real-world drone delivery airport, demonstrating feasibility of mmE-Loc.

\section{System Overview}
The mmE-Loc enhances the mmWave radar with an event camera to achieve accurate and low-latency drone ground localization, allowing the drone to rapidly adjust its location state and perform a precise landing.
Given the critical importance of safety in commercial drone operations, mmE-Loc can work in conjunction with RTK or visual markers to ensure precise landing performance.
In this section, we mathematically introduce the problem that mmE-Loc tries to address and provide an overview of the system design.



\vspace{-0.3cm}
\subsection{Problem Formulation}
In this section, we illustrate key variables in mmE-Loc and introduce the system's inputs and outputs.


\textbf{Reference systems.} \label{3.2}
There are four reference (\aka, coordinate) systems in mmE-Loc: 
$(i)$ the Event camera reference system $\mathtt{E}$; 
$(ii)$ the Radar reference system $\mathtt{R}$; 
$(iii)$ the Object reference system $\mathtt{O}$;
$(iv)$ the Drone reference system $\mathtt{D}$.
Note that a drone can be considered as an object.
For clarity, before an object is identified as a drone, we utilize $\mathtt{O}$. 
Once confirmed as a drone, we use $\mathtt{D}$ for the drone and continue using $\mathtt{O}$ for other objects.
Throughout the operation of system, $\mathtt{E}$ and $\mathtt{R}$ remain stationary and are rigidly attached together, while $\mathtt{O}$ and $\mathtt{D}$ undergo changes in accordance with movement of the object and the drone, respectively. 
The transformation from $\mathtt{R}$ to $\mathtt{E}$ can be readily obtained from calibration \cite{wang2023vital}. 

\textbf{Goal of mmE-Loc.}
The goal of mmE-Loc is to determine 3D location of the drone, defined as $t_{\mathtt{ED}}$, the translation from coordinate system $\mathtt{D}$ to $\mathtt{E}$.
Specifically, mmE-Loc optimizes and reports 3D location of drone $(l_x, l_y, l_z)$ at each timestamp $i$ with input from event stream and radar sample.
$t_{\mathtt{ED}}$ and ($l_x$, $l_y$, $l_z$) are equivalent representations of the drone’s location and can be inter-converted with Rodrigues’ formula \cite{min2021joint}. 
The former representation is adopted in the paper, as it is commonly used in drone flight control systems.

\begin{figure*}[t]
    \setlength{\abovecaptionskip}{0.3cm} 
    \setlength{\belowcaptionskip}{-0.2cm}
    \setlength{\subfigcapskip}{-0.25cm}
    \centering
        \includegraphics[width=1.85\columnwidth]{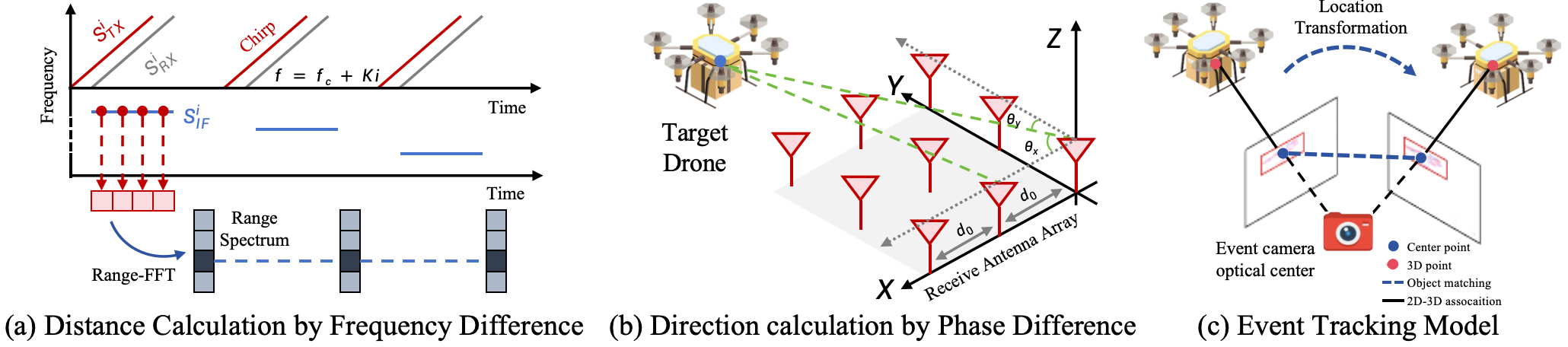}
        \vspace{-0.3cm}
    \caption{Illustration of tracking models in Consistency-instructed Collaborative Tracking algorithm.}
    \label{CCT}
    \vspace{-0.2cm}
\end{figure*}

\vspace{-0.3cm}
\subsection{Overview}
As illustrated in \fig \ref{overview}, mmE-Loc comprises two key modules: 

\noindent $\bullet$ 
The \textit{CCT} (\textbf{C}onsistency-instructed \textbf{C}ollaborative \textbf{T}racking) for noise filtering, drone detection, and preliminary localization of the drone.
This module utilizes time-synchronized event streams and mmWave radar measurements as inputs. 
Subsequently, the \textit{Radar Tracking Model} processes radar measurements to generate a sparse 3D point cloud. 
Meanwhile, the \textit{Event Tracking Model} takes into the stream of asynchronous events for event filtering, drone detection, and tracking. 
Finally, \textit{Consistency-instructed Measurements Filter} aligns the outputs of both tracking models by leveraging \textit{temporal-consistency} between the two modalities. 
It then utilizes the drone's periodic micro-motion to extract drone-specific measurements and achieve drone preliminary localization.

\noindent $\bullet$
The \textit{GAJO} (\textbf{G}raph-informed \textbf{A}daptive \textbf{J}oint \textbf{O}ptimization) for fine localization and trajectory optimization of the drone.
Based on the operational principles of two sensors and their respective noise distributions, \textit{GAJO} incorporates a meticulously designed \textit{factor graph-based optimization} method. 
This module employs the \textit{spatial-complementarity} from both modalities to unleash the potential of event camera and mmWave radar in drone ground localization.
Specifically, \textit{GAJO} jointly fuses the preliminary location estimation from the \textit{Event Tracking Model} and the \textit{Radar Tracking Model} and adaptively refines them, determining the fine location of drone with $ms$-level processing time.
\vspace{-0.3cm}
\section{System Design} \label{4}
In this section, we introduce \textit{CCT}: Consistency-instructed Collaborative Tracking for noise filtering, detection, and preliminary localization of drone (§ \ref{4.1}). 
Subsequently, we delve into \textit{GAJO}: Graph-informed Adaptive Joint Optimization for fine localization and trajectory optimization of drone (§ \ref{4.2}).

\vspace{-0.2cm}
\subsection{\textit{CCT}: Consistency-instructed \\ Collaborative Tracking} \label{4.1}

The mmWave radar is prone to signal multipath effects, leading to inaccurate point cloud data.
Meanwhile, the event camera captures per-pixel brightness changes asynchronously, which are frequently influenced by non-drone factors such as shadows.  
However, the absence of intrinsic drone semantic information, combined with significant differences in dimension and patterns between these two modalities, presents challenges for noise filtering. 
This results in drone detection bottlenecks, which further reduce the efficiency and accuracy of localization.
Therefore, in this part, we focus on enhancing noise filtering and drone detection, while providing preliminary localization of the drone.


To address this challenge, we explore the operational principles of both sensors. 
Our design is based on observations: \textit{(i) Event camera and mmWave radar demonstrate temporal consistency and distinct response mechanisms.}
Event camera and mmWave radar maintain $ms$-level latency.
Additionally, event cameras are unaffected by multipath effects, whereas mmWave radar remains impervious to changes in brightness.
\textit{(ii) Drone exhibits periodic micro motion features (\eg, propeller rotation),} which can serve as stable and distinctive features of drone.
These facilitate efficient cross-modal noise filtering by aligning measurements from the event camera and mmWave radar and enable drone detection by extracting drone measurements through periodic micro-motions.
\begin{figure}[t]
    \setlength{\abovecaptionskip}{0.2cm} 
    \setlength{\belowcaptionskip}{-0.3cm}
    \setlength{\subfigcapskip}{-0.4cm}
    \centering
        \includegraphics[width=0.85\columnwidth]{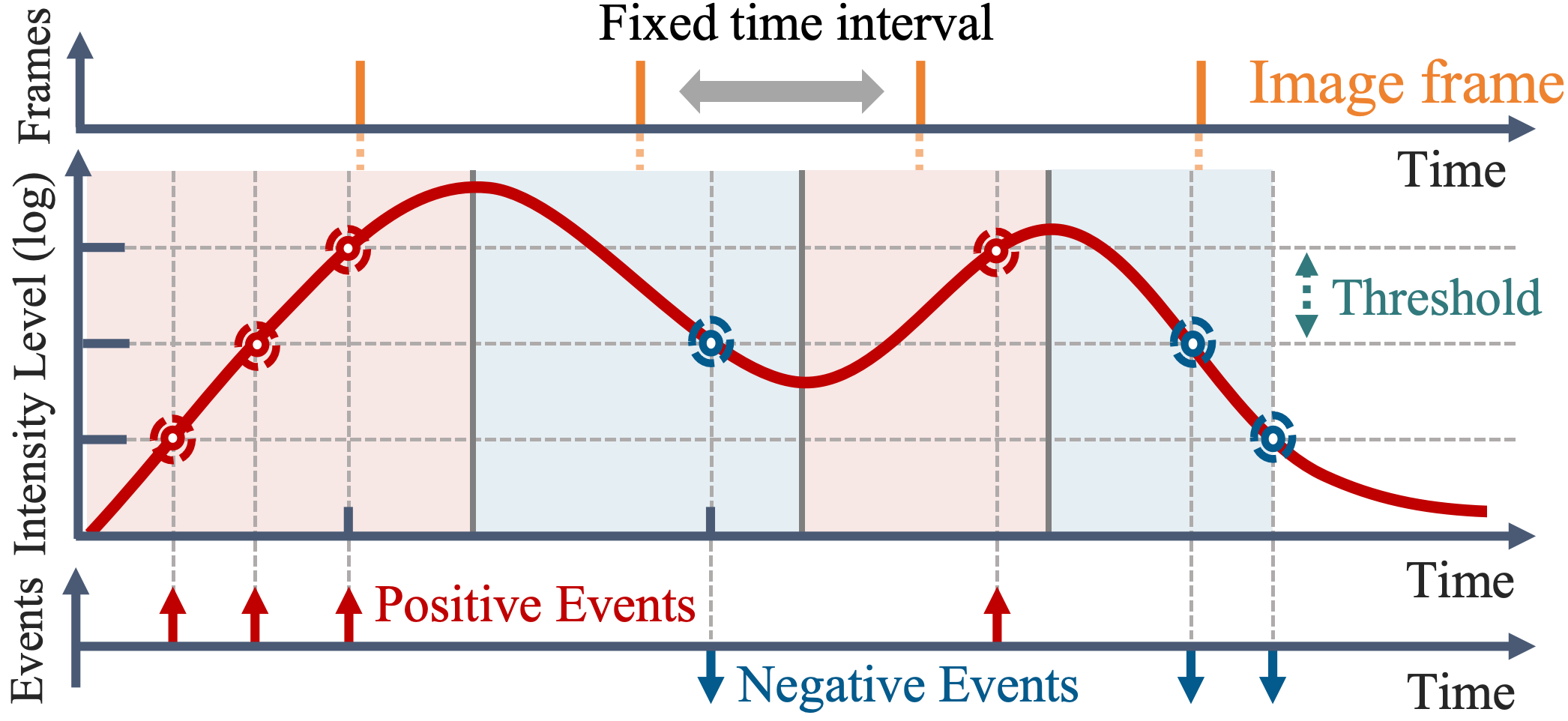}
    \caption{Illustration of synchronous frames and asynchronous events. \textnormal{Frame cameras use a global shutter to capture images at fixed intervals, while each pixel in an event camera responds independently, generating events asynchronously when intensity changes exceed a threshold.}}
    \label{event}
    \vspace{-0.4cm}
\end{figure} 

\begin{figure*}[t]
    \setlength{\abovecaptionskip}{0.4cm} 
    \setlength{\belowcaptionskip}{-0.34cm}
    \setlength{\subfigcapskip}{-0.25cm}
    \centering
        \includegraphics[width=2\columnwidth]{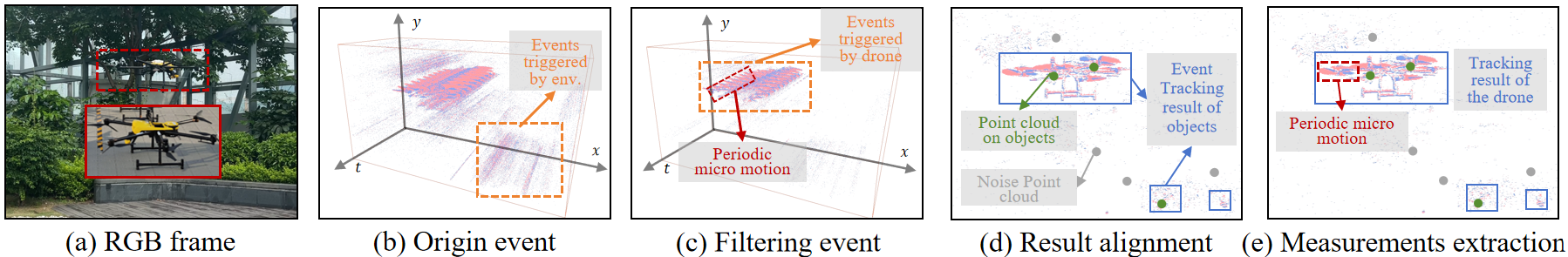}
        \vspace{-0.28cm}
    \caption{Step-by-step filtering performance. \textnormal{The \textit{CCT} module in mmE-Loc eliminate noise events, mmWave point cloud and erroneous detection by employing \textit{temporal-consistency} of both modalities.}}
    \label{performance}
    \vspace{-0.2cm}
\end{figure*}

To realize this idea, we design \textit{CCT}, a lightweight cross-modal drone detector and tracker.
\textit{CCT} includes several components:
$(i)$ a Radar Tracking Model (§\ref{4.1.1}) providing sparse point cloud indicating distance and direction information of objects;
$(ii)$ an Event Tracking Model (§\ref{4.1.2}) for event filtering, detection, and tracking of objects;
$(iii)$ a Consistency-instructed Measurements Filter (§\ref{4.1.3}) utilizes temporal consistency between both modalities and drone's periodic micro-motion features to extract detection and point cloud of drone, facilitating the preliminary localization.

\subsubsection{\textbf{Radar Tracking Model}} \label{4.1.1}
In this part, we calculate the distance $D$ and direction vector $\vec{v}$ between the radar and objects, along with a preliminary estimation of the object's location, as depicted in \fig \ref{CCT}a and \fig \ref{CCT}b.

\textbf{Distance calculation.} 
As shown in \fig \ref{CCT}a, the frequency difference between the transmitted (TX) and received (RX) signals indicates the signal propagation time, revealing the distance between the object and the radar.
Denoting $D^i$ as the distance at time $i$, TX and RX signals as:
\begin{equation}
\begin{aligned}
S_{TX}^i\!=\!\exp \left[j\left(2 \pi f_c i+\pi K i^2\right)\right], 
S_{RX}^i\!=\!\alpha S_{TX}\left[i-2D^i/{c}\right],
\end{aligned}
\end{equation}
where $\alpha$ denotes the attenuation rate, $f_c$ is the initial frequency, $K$ represents the chirp slope of the FMCW signal, and $c$ stands for speed of light.
The TX and RX signals undergo mixing and low-pass filter (LPF) to extract intermediate frequency signal (IF signal) $s(t)$: 
\begin{equation}
S_{IF}^i=LPF(S_{TX}^{i*} S_{RX}^{i}) \approx \alpha \exp \left[j 2 \pi\left(2KD^i/c\right)i\right].
\end{equation}
The frequency value $f_{IF}$ within $S_{IF}^i$ encapsulates distance information. 
After the Range-FFT operation $S_{IF}^i$, $f_{IF}$ is extracted, facilitating distance calculation $D^i=c f_{IF} / 2K$.

\textbf{Direction calculation.}
Using a fixed antenna array, the mmWave radar determines the object's direction by employing two orthogonal linear arrays. 
As depicted in \fig \ref{CCT}b, each linear array captures an Angle of Arrival (AoA), calculated from the phase difference between adjacent antennas spaced apart by $d$ as $cos \theta = \Delta \phi \lambda/2 \pi d$, where $\theta$ represents AoA, $\lambda$ denotes the wavelength and $\Delta \phi$ indicates the phase difference. 
With two orthogonal arrays, the radar obtains two AoAs, $\theta_x$ and $\theta_y$. The unit vector indicating the object's direction at time $i$ is given by
$\vec{v}^i=[\cos \theta_x \cos \theta_y \sqrt{1-\cos ^2 \theta_x-\cos ^2 \theta_y}]^{\mathrm{T}}$.

Using the distance and angle information obtained above, along with the spatial relationship between radar and event camera, we can determine the preliminary 3D location estimation of the object in $\mathtt{E}$ as $P_E = D\vec{v}+t_{ER}$.
We then leverage the mmWave radar for object 3D location tracking, estimating the translation $t_{\mathtt{EO}}$ of the object from $\mathtt{O}$ to $\mathtt{E}$ at time $i$:
\begin{equation}
\begin{aligned}
\vspace{-0.2cm}
t_{\mathtt{EO}}^i & =t_{\mathtt{EO}}^{i-1}+U_{\mathtt{E}}^{i}+w^i + w^{i-1} \\
& =t_{\mathtt{EO}}^{i-1}+\left(P_{\mathtt{E}}^i-P_{\mathtt{E}}^{i-1}\right)+w^i + w^{i-1}.
\vspace{-0.2cm}
\end{aligned}
\end{equation}
$U_{\mathtt{E}}^{i}$ is discrepancy between two radar calculation results at times $i$ and ${i-1}$ in $\mathtt{E}$. $w_i$ and $w_{i-1}$ signify the measurement noise.

\revise{
While mmWave radars excel at estimating object depth along the radial direction, they struggle to accurately capture horizontal and vertical (tangential) motion \cite{qian20203d, zhang2023push}.
To address this issue, we introduce the event camera, which has similar latency but a different sensing principle. 
With high spatial resolution, the event camera detects objects and compensates for mmWave radars' limitations in the tangential direction.
}

\subsubsection{\textbf{Event Tracking Model}} \label{4.1.2}
In this part, we demonstrate the process of noise filtering from a stream of asynchronous events, and how to detect and track objects with the filtered events, as depicted in \fig \ref{CCT}c.
Compared to frame cameras that use a global shutter to capture images at fixed intervals, event cameras record pixel-wise intensity changes with $ms$-level resolution and sampling latency, enabling high-speed motion capture without blurring but adding complexity to noise filtering and object detection (Fig. \ref{event}).


\textbf{Similarity-informed event filtering.}
Event cameras are prone to noise from transistor circuits and other non-idealities, requiring pre-processing filtering. 
For the $i^{th}$ event $e^i_{(x, y)}$ with the timestamp $t^i_{(x, y)}$, we assess the timestamp ($t^i_{n(x, y)}$) of the most recent neighboring event in all directions. 
Events with a time difference less than the threshold $T_n$ are retained, indicating object activity, while those exceeding it are discarded as noise (\fig \ref{performance}b, \fig \ref{performance}c).
\revise{
We utilize the Surface of Active Events (SAE) \cite{lin2020efficient} to manage events, mapping coordinates $(x, y)$ to timestamps $(t_l, t_r)$.
Upon a new event's arrival, $t_l$ updates accordingly, and $t_r$ updates only if the previous event at the same location occurred outside the time window $T_k$ or had a different polarity. 
Events that update value of $t_r$ are retained.
The event stream, segregated by polarity, is processed with distinct SAEs. 
This method ensures precise spatial-temporal representation, reducing events and conserving computational resources.
}

\textbf{Filter-based detection and tracking.} 
We employ a grid-based method to cluster events to facilitate object detection. 
The camera's field of view is partitioned into elementary cells sized $c_w \times c_h$. 
For each cell, we compare the event count within a specified time interval ($c_{\Delta t}$) to an activation threshold $c_{thres}$. 
Cells surpassing $c_{thres}$ are marked as active and connected to form clusters, serving as object detection results, including those generated by the drone.
\revise{
For tracking, we deploy Kalman filter-based trackers with a constant velocity motion model, as the Kalman filter provides low-latency estimates with minimal computational cost.
A tracker predicts the state of the current object and associates it with the input cluster that has the largest Intersection Over the Union area. 
The input cluster corrects tracker state, generating bounding boxes, and effectively tracking moving objects, including the drone.
}


\revise{
Using bounding box proposals and the pinhole camera model with projection function $\pi$, we estimate the preliminary 3D locations of objects.
Specifically, the projection function $\pi$ transforms a 3D point $\textbf{X}_\mathtt{E}$ in $\mathtt{E}$ into a 2D pixel $x$ in the image plane as: 
\begin{equation}
x\!=\!\pi\left(\textbf{X}_\mathtt{E}\right)\!=\![f_x X_\mathtt{E} / Z_\mathtt{E}+c_x,
f_y Y_\mathtt{E} / Z_\mathtt{E}+c_y]^T, 
\textbf{X}_\mathtt{E}\!=\![X_\mathtt{E}, Y_\mathtt{E}, Z_\mathtt{E} ]^T,
\end{equation}
where $[f_x, f_y]^T$ is the focal length of the event camera, and $[c_x, c_y]^T$ denotes the principal point, both being intrinsic camera parameters. 
Then, the object's preliminary location at time $i$ is estimated using the center point of bounding box proposal $x^i$ as: 
\begin{equation}
x^i =\pi(\textbf{X}_\mathtt{E}^i)+v^i =\pi(\textbf{X}_\mathtt{O}^i+t_{\mathtt{EO}}^i)+v^i,
\end{equation}
where $\textbf{X}_\mathtt{O}^i$ represents the corresponding 3D point of center point $x^i$ in the object reference $\mathtt{O}$, $v^i$ denotes the random noise.
When extracting center points from bounding box proposals, we first undistort their coordinates. 
}

\vspace{-0.8cm}
\subsubsection{\textbf{Consistency-instructed Measurements Filter}} \label{4.1.3}


The \textit{Event Tracking Model} detects drones and other objects causing light changes, such as indicator lights or shadows. The system must distinguish the landing drone from these objects. Similarly, the \textit{Radar Tracking Model} outputs a 3D point cloud containing both the drone and noise from multipath effects, requiring extraction of the drone’s relevant points.

\textbf{Consistency-instructed alignment.} 
Utilizing the \textit{temporal-consistency} from the event camera and radar, and their distinct mechanisms respond to dynamic objects, we filter event camera results affected by lighting variations on stationary objects and vice versa for radar points influenced by multipath effects.
Specifically, we align synchronized radar points (\textit{Radar Tracking Model}) to each event bounding box (\textit{Event Tracking Model}) (\fig \ref{performance}d). 
Using event camera's projection, we determine that object's location lies along the ray from the camera's optical center through bounding box center. 
The system then identifies the nearest radar points along this ray to isolate the object-associated points.
If no radar point is detected, the bounding box is treated as noise and disregarded.


\textbf{Periodic micro motion-aid measurements extraction.} 
Since each platform supports one drone landing at a time, we need to identify a distinguishing feature of the landing drone, which effectively differentiates the drone from noise, and use it to extract landing drone-specific measurements from the aligned tracking results. 
Our finding is that drones exhibit periodic micro-motions (\eg, propeller rotation), which can serve as stable and distinctive features of the drone. 
We transform the spatio-temporal distribution of events into a heatmap and apply statistical metrics to isolate drone measurements leveraging this feature. 
Specifically, within a time window $[i, i + \delta i]$, events are binned into a 2D histogram where each bin corresponds to a spatial region (\eg., $5\times 5$ pixels). 
Bins containing propeller rotation tend to accumulate more events due to rapid light intensity changes. 
Meanwhile, these propeller rotations generate bipolar events within a bin, while background motion and noise typically result in unipolar events (\eg, from flying birds).
Therefore, we select bins with propeller rotation based on event counts and the proportion of positive events, favoring those with higher counts and a more balanced ratio. 
Finally, we identify event tracking results with the most bins indicative of propeller rotation and corresponding point clouds ($t_{EO}$), designating them as drone tracking results ($t_{ED}$) for preliminary localization from two models as shown in \fig \ref{performance}e.
When multiple drones are scheduled to land, they descend and land sequentially. 
This method accurately identifies the landing drone and extracts relevant measurements.

\vspace{-0.2cm}
\subsection{\textit{GAJO}: Graph-informed Adaptive \\ Joint Optimization} \label{4.2}

The preliminary drone location estimations from the event and radar tracking models suffer from biases. 
Specifically, event camera estimations face scale uncertainty, while radar estimations struggle with limited spatial resolution, scatter center drift, and accumulating drift. 
Additionally, estimations from different models are heterogeneous in precision, scale, and density, complicating the fusion and optimization.
Therefore, in this part, we prioritize accurate drone ground localization and trajectory tracking.

\begin{figure}[t]
    \setlength{\belowcaptionskip}{-0.2cm}
    \setlength{\subfigcapskip}{-0.6cm}
    \centering
        \includegraphics[width=0.95\columnwidth]{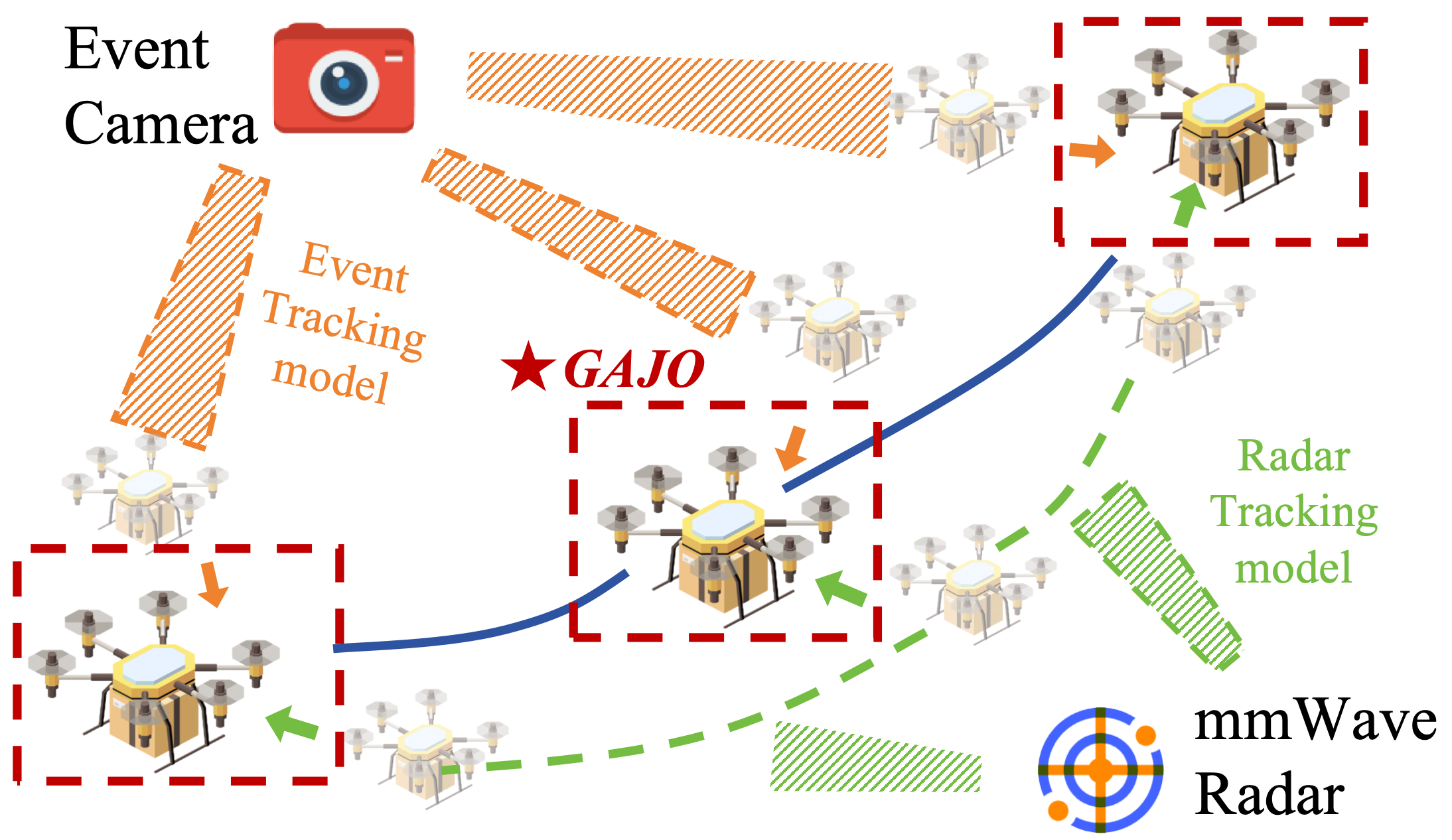}
        \vspace{-0.4cm}
    \caption{Illustration of relationship between \textit{GAJO} and \textit{CCT}. \textnormal{The \textit{GAJO} module harness the \textit{spatial-complementarity} of both modalities through a join optimization.}}
    \label{relationship}
    \vspace{-0.5cm}
\end{figure}

\revise{

Our design is founded on the insight that \textit{the Event Tracking Model and Radar Tracking Model provide distinct features that are spatial-complementarity to each other.} 
As a result, the 2D imaging capability of event cameras and the depth sensing capability of mmWave radar mutually enhance each other when combined, as demonstrated in \fig \ref{relationship}. 
Since both the event stream and mmWave samples are drone-related, fully leveraging the \textit{spatial- complementarity} of these two modalities through joint optimization offers the potential to significantly improve performance. This leads to a trajectory with reduced bias and minimized cumulative drift.
}

To realize this idea and push the limit of localization accuracy while minimizing latency, we introduce a \textit{GAJO}, a factor graph-based location optimization framework designed for low-latency and accurate drone 3D localization (§\ref{4.2.1}).
\textit{GAJO} includes two parallel tightly coupled modules: $(i)$ short-term (inter-SAE tracking) and $(ii)$ long-term (local location optimization) optimizations, collectively enhancing location tracking precision (§\ref{4.2.3}).
Beyond the capabilities of \textit{Event Tracking model} and \textit{Radar Tracking model}, \textit{GAJO} assimilates prior knowledge of drone's flight dynamics to refine the trajectory for enhanced smoothness and accuracy (§\ref{4.2.2}).

\subsubsection{\textbf{Factor graph-based optimization}}\label{4.2.1}

A factor graph comprises variable nodes, indicating the states to be optimized (\eg, $t_{ED}^i$), and factor nodes, representing the probability of certain states given a measurement result. 
In mmE-Loc, measurements are derived from the Event Tracking (ET) model ($x^i$) and Radar Tracking (RT) model ($D^i$, $\vec{v}^i$, and $U_E^{i}$).
To estimate the values of a set of variable nodes $\boldsymbol{\mathcal{X}} = \{t_{ED}^i | i \in \mathcal{T}\}$ given measurements $\boldsymbol{\mathcal{Z}} = \{x^i, D^i, \vec{v}^i, U_E^{i} | i \in \mathcal{T}\}$, \textit{GAJO} optimizes all connected factor nodes based on maximum a posteriori estimation:
\begin{align}
\begin{split}
\hat{\boldsymbol{\mathcal{X}}} & =\underset{\boldsymbol{\mathcal{X}}}{\arg \max } \ p(\boldsymbol{\mathcal{X}} \mid \boldsymbol{\mathcal{Z}}) = \underset{\boldsymbol{\mathcal{X}}} {\arg \max } \  p(\boldsymbol{\mathcal{X}}) \ p(\boldsymbol{\mathcal{Z}} \mid \boldsymbol{\mathcal{X}}) \\
& =\underset{\boldsymbol{\mathcal{X}}}{\arg \max } \ 
p(\boldsymbol{\mathcal{X}}) \prod_{i \in \mathcal{T}} \ p\left(x^i \mid t_{ED}^i\right) p\left(D^i, \vec{v}^i, U_E^{i} \mid t_{ED}^i\right),
\end{split}
\label{factor_graph}
\end{align}
which follows the Bayes theorem.
$p(\boldsymbol{\mathcal{X}})$ is the prior information over $\boldsymbol{\mathcal{X}} = \{t_{ED}^i | i \in \mathcal{T}\}$, which is inferred from drone flight characteristics.
The $p\left(x^i \mid t_{ED}^i\right)$ is the likelihood of the ET model measurements. The $p\left(D^i \mid t_{ED}^i\right)$, $p\left(\vec{v}^i \mid t_{ED}^i\right)$ and $p\left(U_E^{i} \mid t_{ED}^i\right)$ are likelihood of the RT model measurements.

\begin{figure*}[t]
    \setlength{\abovecaptionskip}{0.05cm} 
    \setlength{\belowcaptionskip}{-0.3cm}
    \setlength{\subfigcapskip}{-0.25cm}
    \centering
        \includegraphics[width=1.92\columnwidth]{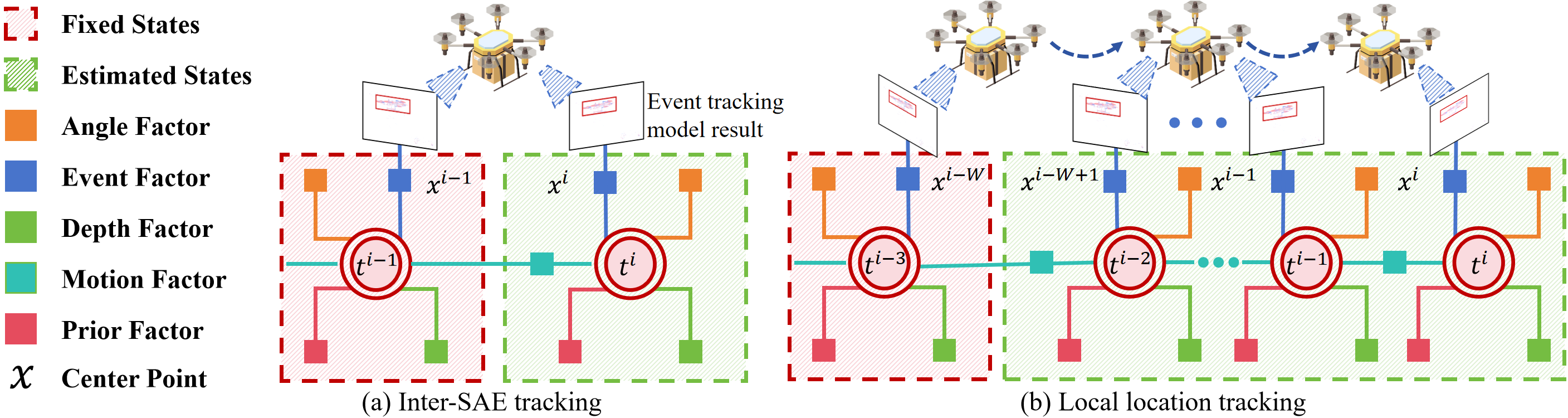}
    \caption{Long-short term optimization based on the factor graph.}
    \label{factorgraph}
    \vspace{-0.25cm}
\end{figure*}

\subsubsection{\textbf{Probabilistic Representation}} \label{4.2.2}
Inferring the drone's location requires prior term and likelihood term in \eqn \eqref{factor_graph}.

\textbf{Prior term.} 
The prior term, $p(t_{ED}^i)$, represents the drone's location probability distribution at time $i$ unaffected by current measurements. Derived from a constant velocity model, it suggests the drone maintains steady speed over short intervals, allowing us to predict the prior location using:
\begin{equation}
\vspace{-0.1cm}
\bar{t}_{\mathrm{ED}}^i-t_{\mathrm{ED}}^{i-1}=t_{\mathrm{ED}}^{i-1}-t_{\mathrm{ED}}^{i-2}.
\end{equation}

\textbf{ET model likelihood.} 
The likelihood $p(x^i|t_{ED}^i)$ from ET model represents the center point distribution at a given drone location. 
In many tracking systems \cite{campos2021orb}, center point noise $v^i$ is assumed Gaussian, proving effective. Thus, likelihood of ET model is:
\begin{equation}
p(x^i|t_{ED}^i) \sim \mathcal{N}(\pi(\textbf{X}_E^i), \sigma_{ET}),
\end{equation}
where $\sigma_{ET}$ is the center point standard deviation.

\textbf{RT model likelihood.}
The likelihood of the RT model $p(D^i \mid t_{ED}^i)$, $p(\vec{v}^i \mid t_{ED}^i)$ and $p(U_E^{i} \mid t_{ED}^i)$ indicates the distribution of the measured distance, angle, and motion at a given drone location.
The distance, angle, and motion from RT model likelihood are:
\begin{equation}
\begin{aligned}
p(D^i \mid t_{ED}^i&) \sim  \mathcal{N}(||t_{ED}^i||, \sigma_{D}), \quad p(\vec{v}^i \mid t_{ED}^i) \sim \mathcal{N}(\vec{v}_{t_{ED}^i}, \sigma_{\vec{v}}), \\
& p(U_E^{i} \mid t_{ED}^i) \sim \mathcal{N}(t_{ED}^i - t_{ED}^{i - 1}, \sigma_{U_E}),
\vspace{-0.4cm}
\end{aligned}
\end{equation}
where $\sigma_{D}$, $\sigma_{\vec{v}}$ and $\sigma_{U_E^{i}}$ are the standard deviation of distance, angle, and motion measurements respectively.

\subsubsection{\textbf{Fusion-based Tracking}} \label{4.2.3}
\revise{
In mmE-Loc, two fusion schemes are employed for sensor fusion and optimization, as depicted in \fig \ref{factorgraph}. 
The first, inter-SAE tracking, aims for instant drone location estimation by minimizing errors across different tracking models simultaneously.  
The second, local location optimization, enhances overall trajectory accuracy through the joint optimization of a selected set of locations.
}

\textbf{Inter-SAE tracking.}
Once the measurements of ET model and RT model $(x^i, D^i, \vec{v}^i, U_E^i)$ received, the prior factor, ET factor and the RT factor are formulated as follows:
\begin{equation}
\begin{aligned}
E^i_{\text {Prior }} & =-\log p\left(t_{ED}^i\right) \propto \left\|t_{ED}^i-\bar{t}_{ED}^i\right\|_{\sigma_{t_{ED}}}^2, \\
E^i_{\mathrm{ET}} & =-\log p\left(x^i \mid t_{ED}^i\right) \propto \rho(\left\| x^i - \pi(\textbf{X}_E^i) \right\|^2_{\Sigma_E}), \\
E^i_{\mathrm{RT}} & =-\log p\left(D^i, \vec{v}^i, U_E^i \mid t_{ED}^i\right) \\
\propto & \left\| ||t_{ED}^i|| \!-\! D^i \right\|^2_{\sigma_D} \!+\!  \left\| \vec{v}_{t_{ED}^i} \!-\! \vec{v}^i \right\|^2_{\sigma_{\vec{v}}} \!+\! \left\| (t_{ED}^i - t_{ED}^{i-1}) \!-\! U_E^i \right\|^2_{\sigma_{U_E}},
\end{aligned}
\end{equation}
where $\left\|e \right\|^2_{\Sigma_E}=e^T\Sigma^{-1} e$.
The symbol $\Sigma_E$ represents the covariance matrix associated with the event camera measurements.

On this basis, the inter-SAE tracking in \fig \ref{factorgraph}a is performed to give an instant location result based on \eqn \eqref{factor_graph} as follows:
\begin{equation}
\begin{aligned}
& \hat{t}_{ED}^i \!=\! \underset{\boldsymbol{t_{ED}^i}}{\arg \max } \ p ( t_{ED}^i \!\mid\! t_{ED}^{i-1}, t_{ED}^{i-2} ) p(x^i \!\mid\! t_{ED}^i) \ p(D^i, \vec{v}^i, U_E^{i} \!\mid\! t_{ED}^i) \\
\vspace{1ex}
& = \underset{\boldsymbol{t_{ED}^i}}{\arg \min } \!-\!\log \!\left(p ( t_{ED}^i \!\mid\! t_{ED}^{i-1}, t_{ED}^{i-2} ) p(x^i \!\mid\! t_{ED}^i) p(D^i\!,\! \vec{v}^i\!,\! U_E^{i} \!\mid\! t_{ED}^i)\right) \\
\vspace{1ex}
& = \underset{\boldsymbol{t_{ED}^i}}{\arg \min } \left( E^i_{\text {prior }} + E^i_{\mathrm{ET}} + E^i_{\mathrm{RT}}\right).
\end{aligned}
\label{inter_frame}
\end{equation}

\textbf{Local location optimization.}
To mitigate cumulative drift, periodic local location optimization is conducted, correcting estimated locations based on multiple consecutive SAEs. 
This optimization entails jointly optimizing the locations of a SAE set denoted as $\mathcal{X}=\underset{i \in \mathcal{T}}{\bigcup}\left\{t_{ED}^i\right\}$, as shown in \fig \ref{factorgraph}b, where $W=|\mathcal{T}|$.
The optimization problem formulation is as follows:
\begin{equation}
\begin{aligned}
\hat{\boldsymbol{\mathcal{X}}} & =\underset{\boldsymbol{\mathcal{X}}}{\arg \max } \ p(\boldsymbol{\mathcal{X}}) \prod_{i \in \mathcal{T}} \ p\left(x^i \mid t_{ED}^i\right) p\left(D^i, \vec{v}^i, U_E^{i} \mid t_{ED}^i\right), \\
& = \underset{\boldsymbol{\mathcal{X}}}{\arg \min } \sum_{i \in \mathcal{T}}\left(E_i^{\mathrm{prior}}+E_i^{\mathrm{ET}}+E_i^{\mathrm{RT}}\right) .
\end{aligned}
\label{local_location}
\end{equation}
It is worth noting that $(i)$ when the local location optimization is triggered, $(ii)$ what is the size of $\mathcal{T}$ ($W$ = $|\mathcal{T}|$), and $(iii)$ how to solve the inter-SAE tracking and local location optimization problems affect the latency and accuracy of localization.
Hence, we enhance the efficiency of \textit{GAJO} through an adaptive optimization method.

\section{Implementation} \label{I}
\subsection{Push the Limit of Accuracy and Latency} \label{I-1}
\textbf{Factor graph solving.}
We represent the estimation problems \eqn \eqref{inter_frame} and \eqn \eqref{local_location} using a factor graph model. To solve the nonlinear least-squares problems, we linearize the observation model and solve the least-squares formulation:
\begin{equation}
\vspace{-0.1cm}
\hat{\boldsymbol{\mathcal{X}}}=\arg \min _{\boldsymbol{\mathcal{X}}}\|A \boldsymbol{\mathcal{X}}-\mathbf{b}\|^2,
\end{equation}
where $A \in \mathbb{R}^{m \times n}$ is the measurement Jacobian and $\mathbf{b} \in \mathbb{R}^m$ is the right-hand side vector. We then utilize QR matrix factorization \cite{bischof1998computing} as $A = Q[R, 0]^T$ and solve the least squares problem $R \hat{\boldsymbol{\mathcal{X}}}=\mathbf{d}$ through backsubstitution to obtain optimized locations $\hat{\boldsymbol{\mathcal{X}}}$, where $R \in \mathbb{R}^{n \times n}$ is the upper triangular square root information matrix, $Q \in \mathbb{R}^{m \times m}$ is an orthogonal matrix, and $\textbf{d} \in \mathbb{R}^n$ \cite{kaess2008isam}. 
Although re-linearizing and regenerating \( R \) with new measurements can help mitigate errors, applying this approach to problem \eqref{local_location} can be computationally expensive, as it requires frequent updates and increased processing overhead, limiting real-time performance.


\begin{algorithm}

\caption{Adaptively Optimization method}
\setlength{\abovedisplayskip}{3pt}
\setlength{\belowdisplayskip}{-1cm}
\label{algorithm}
\KwData{Original factor graph $G$; New measurements $D^i, \vec{v}^i, U^i_E$; square root information matrix $R$}
\KwResult{Updated locations $\hat{\mathcal{X}}$}
$G \leftarrow \mathtt{AddFactorToGraph}(G^i, D^i, \vec{v}, U^i_E)$\;
$R \leftarrow \mathtt{\textbf{IncrementalUpdate}}(G)$\;
$\hat{\mathcal{X}} \leftarrow \mathtt{Backsubstitution}(R)$\;
$L \leftarrow \emptyset$;  $\quad\quad \triangleright \textit{Set of nodes need to be linearized}$\;
\For{all $t^i_{ED} \in \mathcal{X}$ and all $\hat{t}^i_{ED} \in \hat{\mathcal{X}}$}
{
\If{$\hat{t}^i_{ED} - t^i_{ED} \geq \delta$}
{$L \leftarrow L \cup t^i_{ED}$\;}
}
\If{$|L| \geq L_T$ or $||\hat{\mathcal{X}} - \mathcal{X} || \geq \Delta$}
{
\For{all $t^i_{ED} \in \mathcal{X}$}
{
$\mathtt{UpdateLinearizationPoint}(t^i_{ED})$\;
}
$R \leftarrow \mathtt{\textbf{FullUpdate}}(G)$\;
$\hat{\mathcal{X}} \leftarrow \mathtt{Backsubstitution}(R)$\;
}
\end{algorithm}

\vspace{-0.2cm}
\textbf{Adaptive optimization method.}
To tackle this, we propose an adaptive optimization method, leveraging the insight that \textit{new measurements mainly impact localized areas, leaving distant parts unchanged}. 
This allows us to incrementally update $R$ during local optimization \cite{kaess2012isam2}. 
By adaptively combining updated and regenerated $R$, this method reduces latency and enhances accuracy.


\alg \ref{algorithm} shows how adaptively optimization method solves local location optimization problem.
Lines 1-3 depict tracking with incrementally updated $R$, while lines 4-16 show tracking with re-generated $R$. 
Specifically, when receiving new measurements, function $\mathtt{AddFactorToGraph}$ updates the factor graph, and the function $\mathtt{IncrementalUpdate}$ incrementally updates $R$ with new measurements \cite{kaess2012isam2}. 
We then solve local location tracking with this incrementally updated $R$.
When one of two conditions is met, we solve local location tracking with re-generated $R$: 
$(i)$ we track locations that have changed significantly in a set $L = \{t_{ED}^i: \hat{t^i_{ED}} - t^i_{ED} \geq \delta\}$. If enough locations have undergone significant changes (\ie, $|L| \geq L_T$), we solve location tracking with re-generated $R$ output by function $\mathtt{FullUpdate}$.
$(ii)$ The norm of total location changes exceeds a threshold $\Delta$ (\ie, $||\hat{\mathcal{X}} - \mathcal{X} || \geq \Delta$).
Since the local location tracking involves repeatedly solving linear equations, this condition keeps current solution from diverging too far from optimal solution.

\vspace{-0.2cm}
\subsection{Platform Implementation}
\begin{figure*}[t]
    \setlength{\abovecaptionskip}{0.1cm} 
    \setlength{\belowcaptionskip}{-0.5cm}
    \setlength{\subfigcapskip}{-0.25cm}
    \centering
        \includegraphics[width=2\columnwidth]{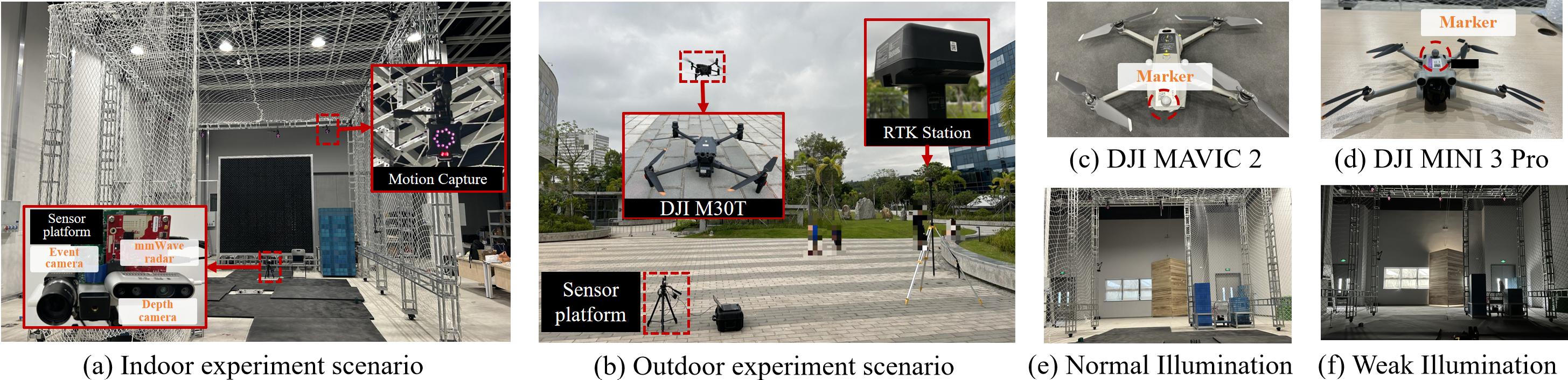}
    \caption{Experimental setup and scenarios of mmE-Loc. \textnormal{(a) A laboratory scenario with motion capture system for ground-truth collection. (b) An outdoor scenario with RTK system for ground-truth collection.  (c)-(f) Different drone with different size (DJI MAVIC 2 and DJI MINI 3 Pro), and different illumination of the laboratory (normal and weak), respectively.}}
    \label{setup}
    \vspace{-0.2cm}
\end{figure*}

\revise{
In this section, we will detail the platform implementation.

\textbf{Sensor platform configuration.}
As illustrated in \fig \ref{setup}, we implement our sensing platform with multiple sensors including 
$(i)$ A Prophesee EVK4 HD evaluation kit, featuring the IMX636ES event-based vision sensor for HD event data (1280 $\times$ 720 pixels) with 47.0\degree  FoV.
$(ii)$ A Texas Instruments (TI) IWR1843 board for transmitting and receiving mmWave signals within the frequency range of 76 $GHz$ to 81 $GHz$ with three transmitting antennas and four receiving antennas.
These antennas are arranged in two linear configurations on the horizontal plane.
$(iii)$ An Intel D435i Depth camera for RGB image capture used in the baseline method.

\textbf{Deployment detail.}
During the experiments, the sensor platform is deployed at the center of the landing pad. 
In the case study at a real-world airport, the platform is positioned at the edge of the landing pad for safety reasons, with the distance between the sensor platform and the center of the pad being less than 1$m$. 
After takeoff, the drone appears within FoV of the event camera. 
All sensors are synchronized through the Robot Operating System (ROS). 
mmE-Loc is deployed on a PC with Ubuntu 20.04, featuring an Intel i7-12900K CPU, 32GB of RAM, and an NVIDIA GeForce GTX 1070 GPU. 
For practical deployments, the sensor platform is expected to be positioned at the center of the landing pad, with its height aligned to that of that of the pad, ensuring the drone remains within the event camera’s FoV at all times for continuous localization.

}
\vspace{-0.3cm}
\section{Evaluation} \label{6}


\begin{figure*}
\setlength{\abovecaptionskip}{-0.1cm} 
\setlength{\belowcaptionskip}{-0.2cm}
\setlength{\subfigcapskip}{-0.3cm}
    \centering
        \subfigure[Indoor Accuracy Comparison]{
            \centering
            \includegraphics[width=0.66\columnwidth]{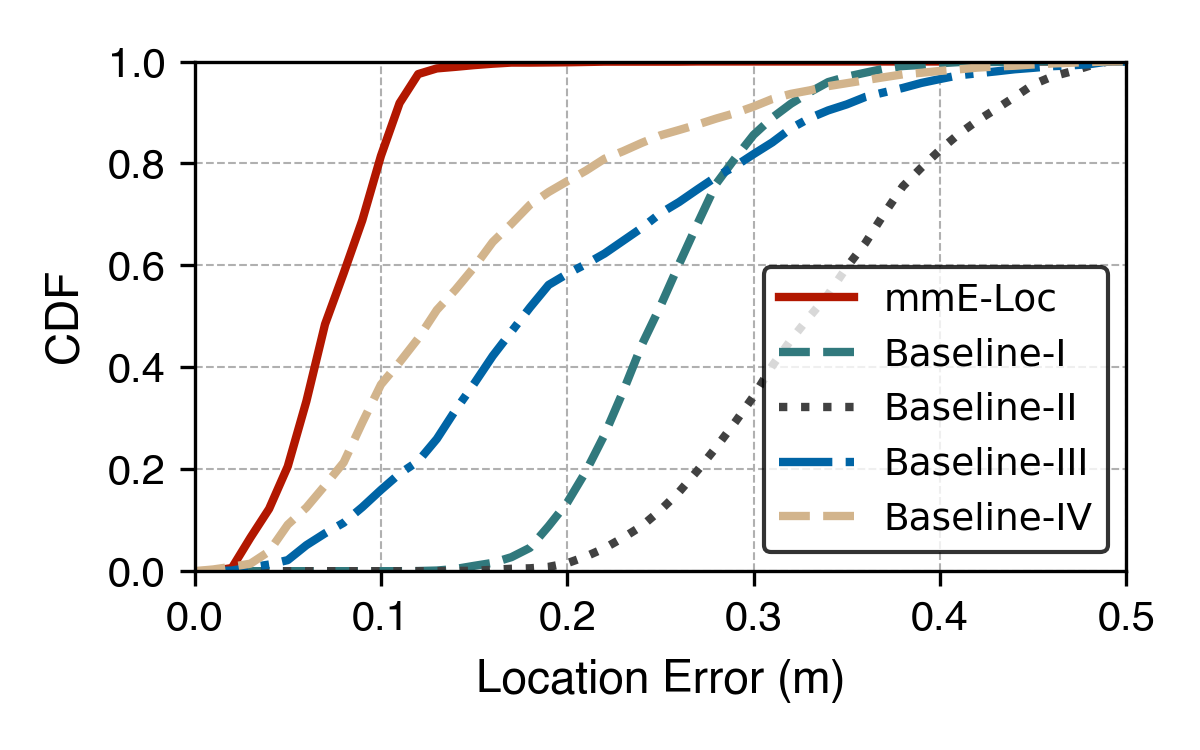}
            \label{fig:indoor_cdf}
        }%
        \subfigure[Outdoor Accuracy Comparison]{
            \centering
            \includegraphics[width=0.66\columnwidth]{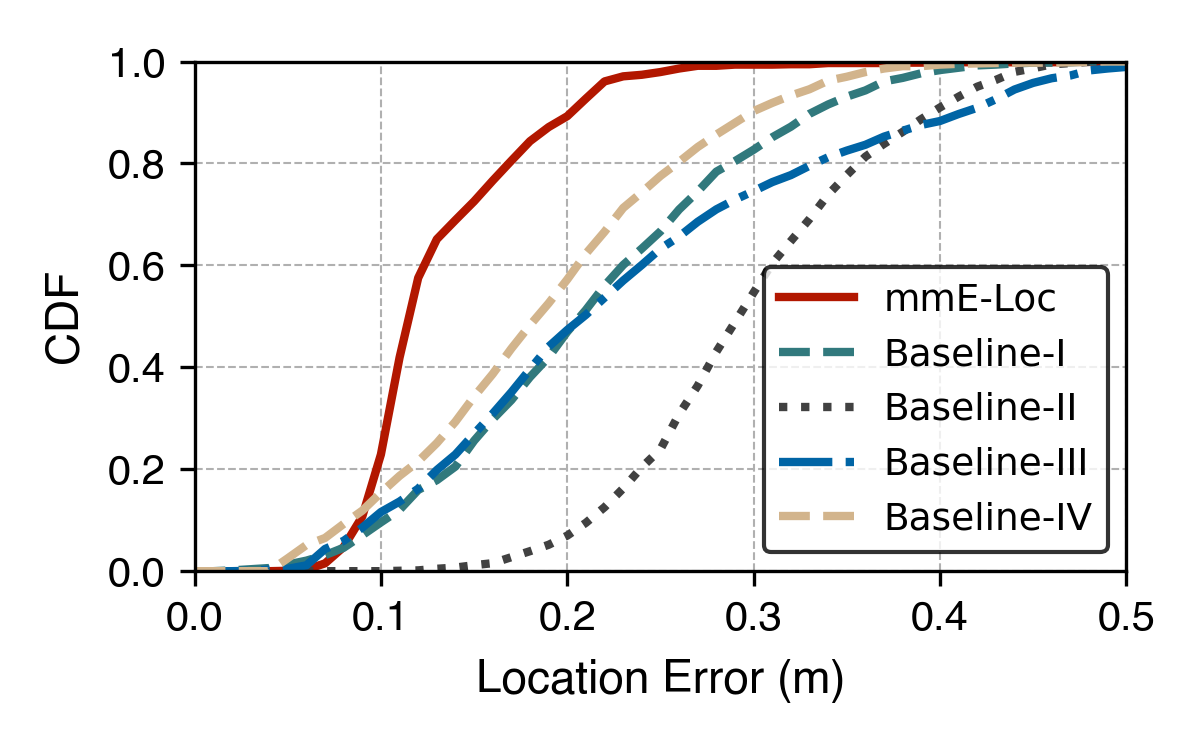}
            \label{fig:outdoor_cdf}
        }%
        \subfigure[Latency Comparison]{
            \centering
            \includegraphics[width=0.66\columnwidth]{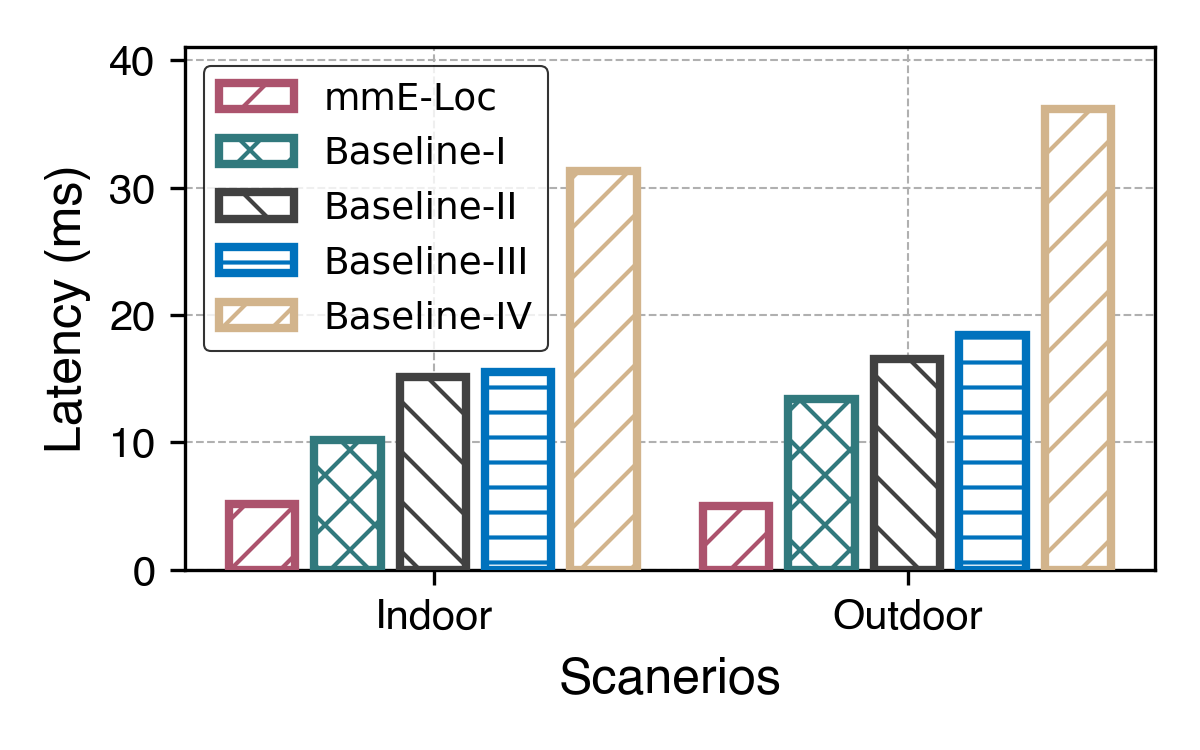}
            \label{fig:indoor_runtime}
        }%
    \caption{Overall performance comparison of mmE-Loc and four related works.}
    \label{fig:Overallperf}
    \vspace{-0.35cm}
\end{figure*}


\begin{figure}
\setlength{\abovecaptionskip}{-0.15cm} 
\setlength{\belowcaptionskip}{-0.2cm}
\setlength{\subfigcapskip}{-0.25cm}
  \begin{minipage}[t]{0.48\columnwidth}
    \centering
    \includegraphics[width=1\columnwidth]{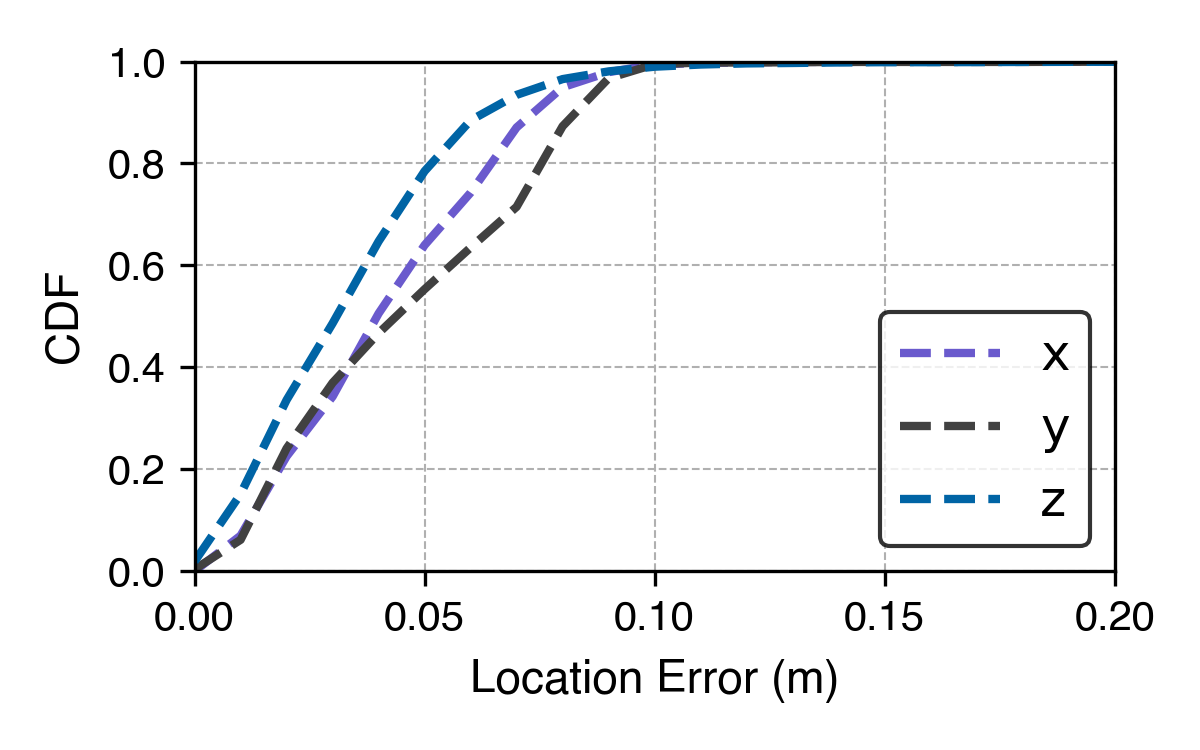}
    \caption{Error distribution along the x, y, and z.}
    \label{fig:xyz}
  \end{minipage}
  \begin{minipage}[t]{0.48\columnwidth}
    \centering
    \includegraphics[width=1\columnwidth]{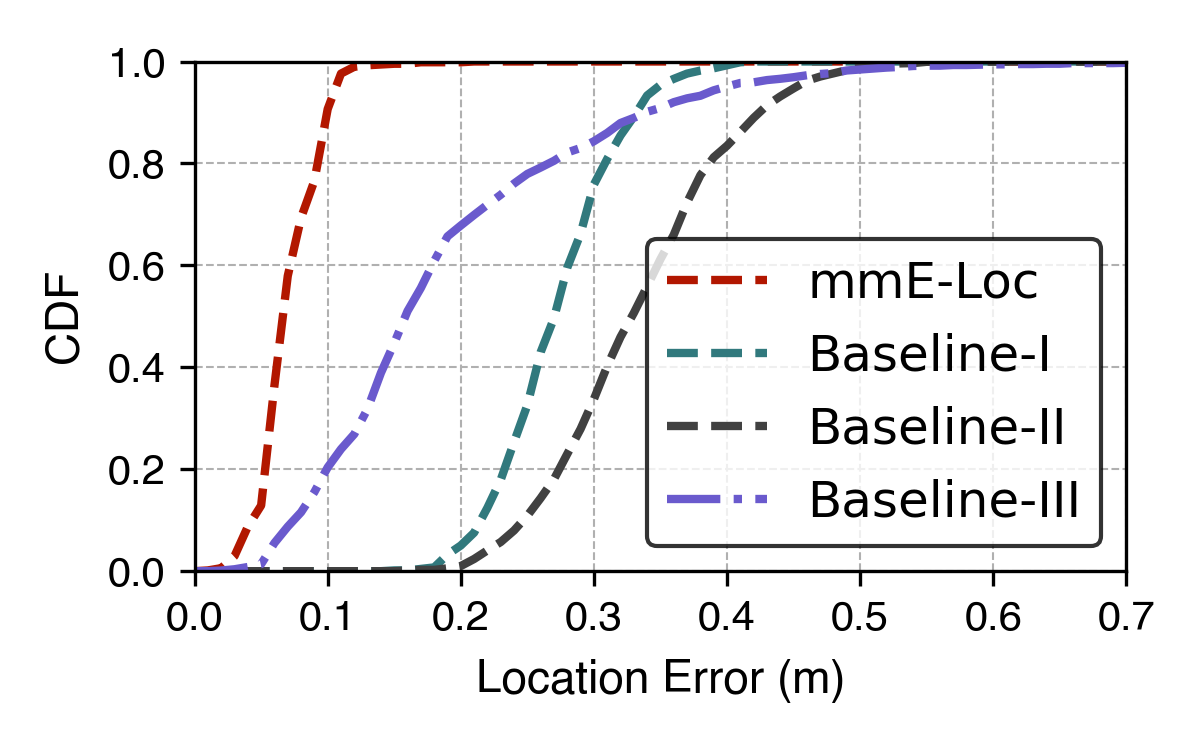}
    \caption{Error in the near-distance setting}
    \label{fig:near_cdf}
  \end{minipage}
  \hfill
  \vspace{-0.5cm}
\end{figure}

Our evaluation of mmE-Loc is comprehensive and grounded entirely in real-world experimentation.  
We begin with experimental settings in §\ref{5.1}. 
Then, in §\ref{5.2}, we focus on key performance metrics: localization accuracy and latency. 
§\ref{5.3} delves into external factors impacting mmE-Loc, such as drone characteristics and environmental conditions. 
§\ref{5.4} assesses benefits of sensor fusion and the performance of various modules. 
Finally, in §\ref{5.5}, we evaluate system load, including latency, CPU usage, and memory consumption.


\vspace{-0.3cm}
\subsection{Experimental Methodology} \label{5.1}
In this section, we will outline the experimental methodology.

\textbf{Experiment setting.} 
We use a deployment setup and drone equipment that closely mimic real-world applications. 
\fig \ref{setup} shows the experimental scenarios in both an \textit{indoor} laboratory and an \textit{outdoor} flight test site. 
The setup in \fig \ref{setup}a includes the event camera and mmWave radar mounted together \textit{at the ground of the experimental area}.
We evaluate mmE-Loc on various target drones:
$(i)$ DJI Mini 3 Pro: 0.25$m$ $\times$ 0.36$m$ $\times$ 0.07$m$ with unfolded propellers.
$(ii)$ DJI MAVIC 2: 0.32$m$ $\times$ 0.24$m$ $\times$ 0.08$m$ with unfolded propellers.
$(iii)$ DJI M30T: 0.49$m$ $\times$ 0.61$m$ $\times$ 0.22$m$ with unfolded propellers.
We conducted extensive experiments over 30 hours, collecting more than 400GB of raw data.


\textbf{Ground truth.}
In the indoor scenario, we used a motion capture system with fourteen cameras. 
This system covers an \textit{8m $\times$ 8m $\times$ 8m} area, providing localization accuracy within \textit{1 mm}, allowing precise performance evaluation under controlled conditions.
In the outdoor scenario,  we conducted experiments at a secluded site with excellent GPS signal reception. We set up a Real-Time Kinematic Positioning (RTK) base station to rebroadcast the GPS signal phase, ensuring high-fidelity RTK processing. The RTK localization results served as the ground truth for our outdoor tests.
%

\textbf{Comparative Methods.}
We compare mmE-Loc under various conditions with three related systems:
$(i)$ \textit{Baseline-I} \cite{zhao20213d}: a SOTA point cloud-based 
drone localization system using single-chip mmWave radar, which is a ground localization system.
$(ii)$ \textit{Baseline-II} \cite{falanga2020dynamic}: a SOTA single event camera-based drone ground localization system applies to drones with known geometries.
The original method, designed for onboard obstacle localization, lacks publicly available code. 
We implement its monocular 3D localization module which uses drone physical characteristics, adapting it for \textit{ground-based} drone localization.
For a fair comparison, we exclude the original method’s ego-motion compensation used for noise reduction due to drone movement.
$(iii)$ \textit{Baseline-III} \cite{falanga2020dynamic}: a SOTA dual event camera-based drone localization system. 
The original method utilizes onboard stereo event cameras for obstacle localization, but its source code is not available. Our implementation excludes the ego-motion compensation component from the original method, implementing its stereo 3D location estimation module to enable \textit{ground-based} drone localization.
$(iii)$ \textit{Baseline-IV} \cite{shuai2021millieye}:
a SOTA deep learning-based object localization system utilizes monocular images and mmWave point clouds as input, harnessing the complementary information from both the radar and the camera. 
We pre-train the neural networks and apply a Kalman Filter to adapt the original method for \textit{ground-based} 3D drone localization.



\revise{
\textbf{Evaluation Metrics.}
The mmE-Loc continuously reports the drone's location estimation with low latency, ensuring accurate and responsive localization. 
Similar to related works, we measure location estimation error in meters and processing latency in milliseconds, allowing for a direct comparison with existing methods in terms of both accuracy and computational efficiency.
}

\textbf{Robustness Experiments.}
During the drone landing, localization may be performed under varying lighting conditions.
To assess the adaptability of mmE-Loc in different scenarios, we conduct experiments across a range of conditions, including different environments, drone models, lighting intensities (\fig \ref{setup}a-f), and background dynamics (controlled by the presence of other moving objects, \eg, balls). 
We also evaluate the impact of varying distances, occlusions (by partially obstructing the drone within the field of view of the event camera), and velocities to highlight the robustness.
In this part, Baseline-IV is excluded from comparison due to its high latency stemming from frame camera's exposure time.

\vspace{-0.4cm}
\subsection{Overall Performance} \label{5.2}
In this section, we demonstrate overall performance of mmE-Loc.

\revise{

\textbf{Drone localization.} 
\fig \ref{fig:Overallperf}a illustrates the localization performance of mmE-Loc compared to four baselines in an indoor environment, using a DJI Mini 3 Pro drone.
mmE-Loc's average end-to-end localization error is 0.083$m$, which outperforms the other baselines with errors of 0.261$m$, 0.345$m$, and 0.209$m$, 0.160$m$, making it suitable for aiding in landing.
\fig \ref{fig:Overallperf}b shows mmE-Loc's localization performance in an outdoor setting using a DJI M30T drone. 
mmE-Loc achieves the lowest average error of 0.135$m$ compared to the other baselines, outperforming them by 39.2\%, 56.0\%, 43.8\%, 31.8\%. 
\fig \ref{fig:xyz} shows error distribution along the x, y, and z dimensions during the typical landing process, and \fig \ref{fig:near_cdf} shows errors in the near-distance setting.
Baseline-I introduces point cloud errors due to phase center offset, causing the point cloud data to deviate from the drone’s geometric center. Additionally, in indoor environments, radar measurements are susceptible to specular reflections, diffraction, and multi-path effects, further contributing to measurement errors.
Baseline-II uses a single event camera with known drone geometries, while Baseline-III, with dual event cameras, estimates depth through stereo vision.
Both are prone to errors when drone detection is affected by outdoor environmental noise (\eg, birds), leading to depth estimation inaccuracies.
Baseline IV incorporates deep learning, which can lead to increased errors when used in environments not present in the training set.
The results demonstrate that mmE-Loc achieves significant improvements by combining complementary strengths of radar and event camera.
mmE-Loc does not require a pre-training procedure, ensuring its applicability.
}

\begin{figure*}
\setlength{\abovecaptionskip}{-0.15cm} 
\setlength{\belowcaptionskip}{-0.2cm}
\setlength{\subfigcapskip}{-0.25cm}
  \begin{minipage}[t]{0.692\columnwidth}
    \centering
    \includegraphics[width=1\columnwidth]{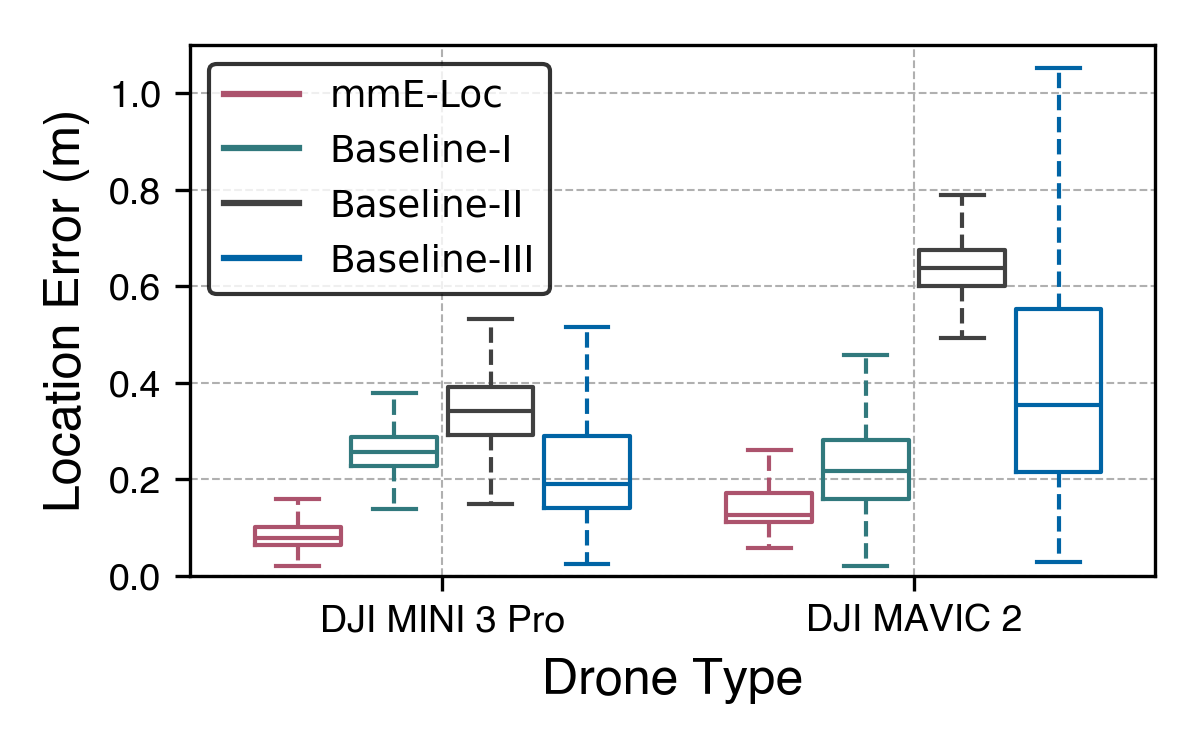}
    \caption{Impact of Drone Type}
    \label{fig:drone_type}
  \end{minipage}
  \begin{minipage}[t]{0.692\columnwidth}
    \centering
    \includegraphics[width=1\columnwidth]{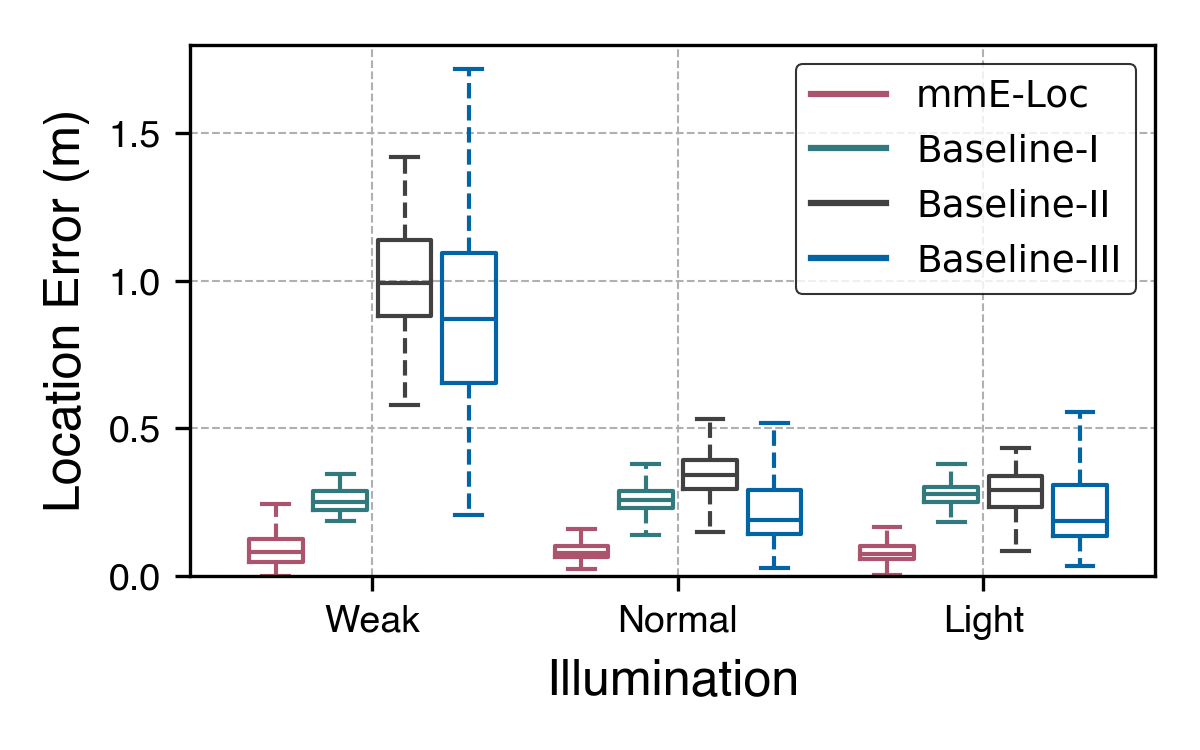}
    \caption{Impact of Env. \& Illu.}
    \label{fig:illumination}
  \end{minipage}
  \begin{minipage}[t]{0.692\columnwidth}
    \centering
    \includegraphics[width=1\columnwidth]{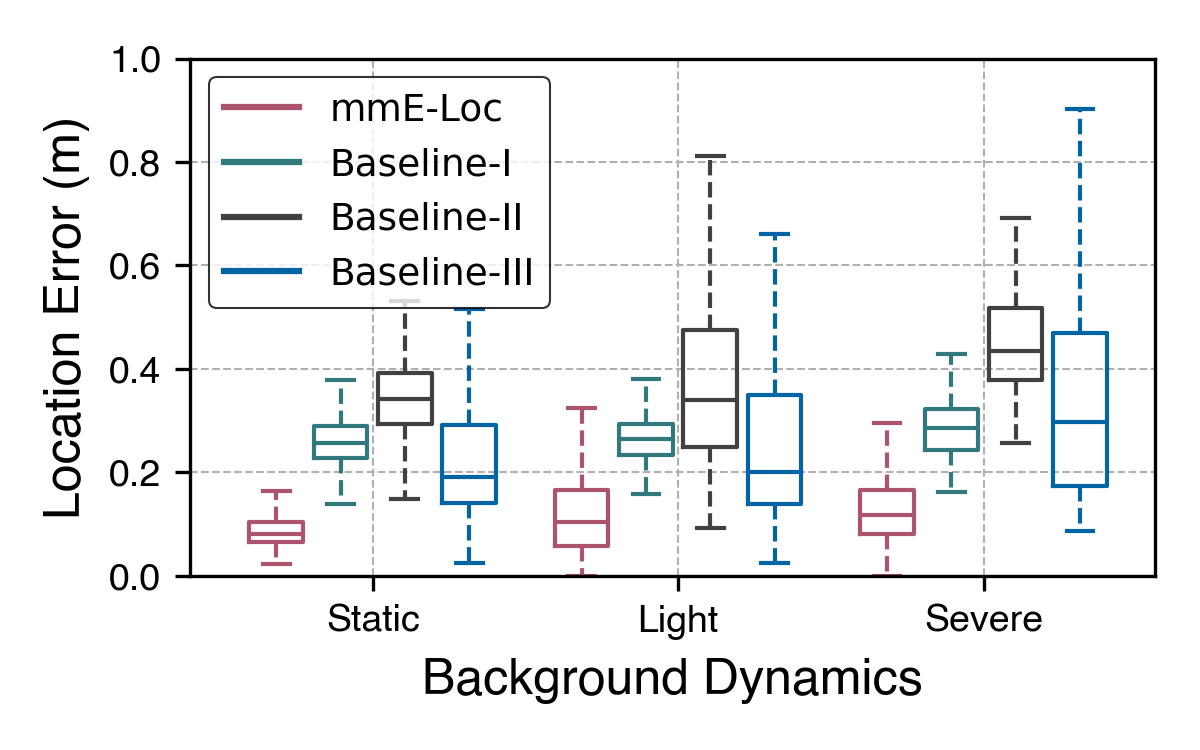}
    \caption{Impact of Background}
    \label{fig:background}
  \end{minipage}
  \hfill
  \vspace{-0.61cm}
\end{figure*}

\begin{figure*}
\setlength{\abovecaptionskip}{-0.15cm} 
\setlength{\belowcaptionskip}{-0.2cm}
\setlength{\subfigcapskip}{-0.25cm}
  \begin{minipage}[t]{0.692\columnwidth}
    \centering
    \includegraphics[width=1\columnwidth]{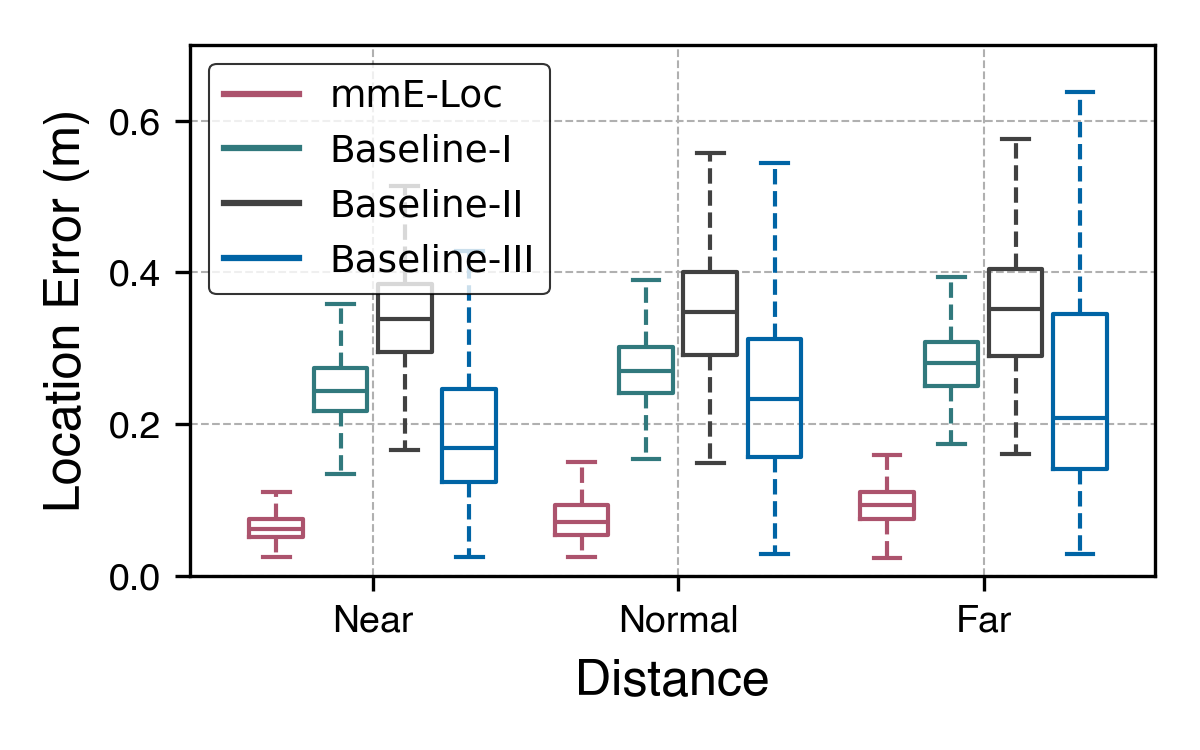}
    \caption{Impact of Distance}
    \label{fig:distance}
  \end{minipage}
  \begin{minipage}[t]{0.692\columnwidth}
    \centering
    \includegraphics[width=1\columnwidth]{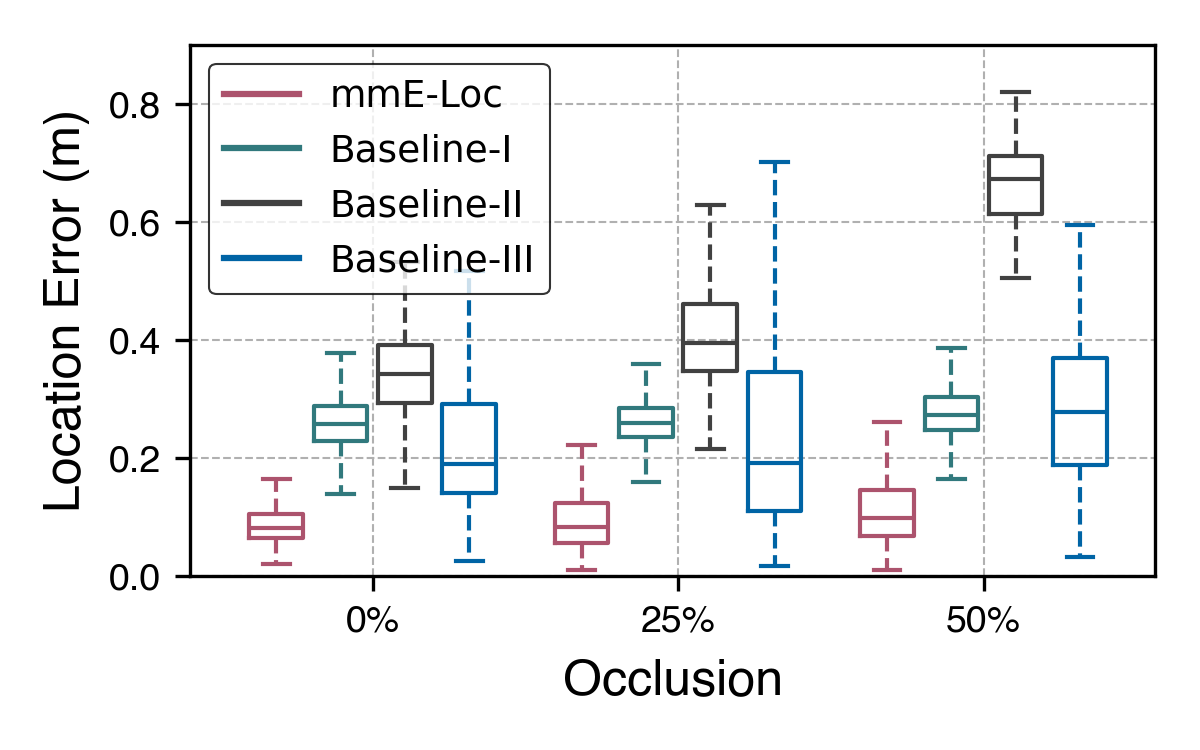}
    \caption{Impact of Occlusion}
    \label{fig:occlusion}
  \end{minipage}
  \begin{minipage}[t]{0.692\columnwidth}
    \centering
    \includegraphics[width=1\columnwidth]{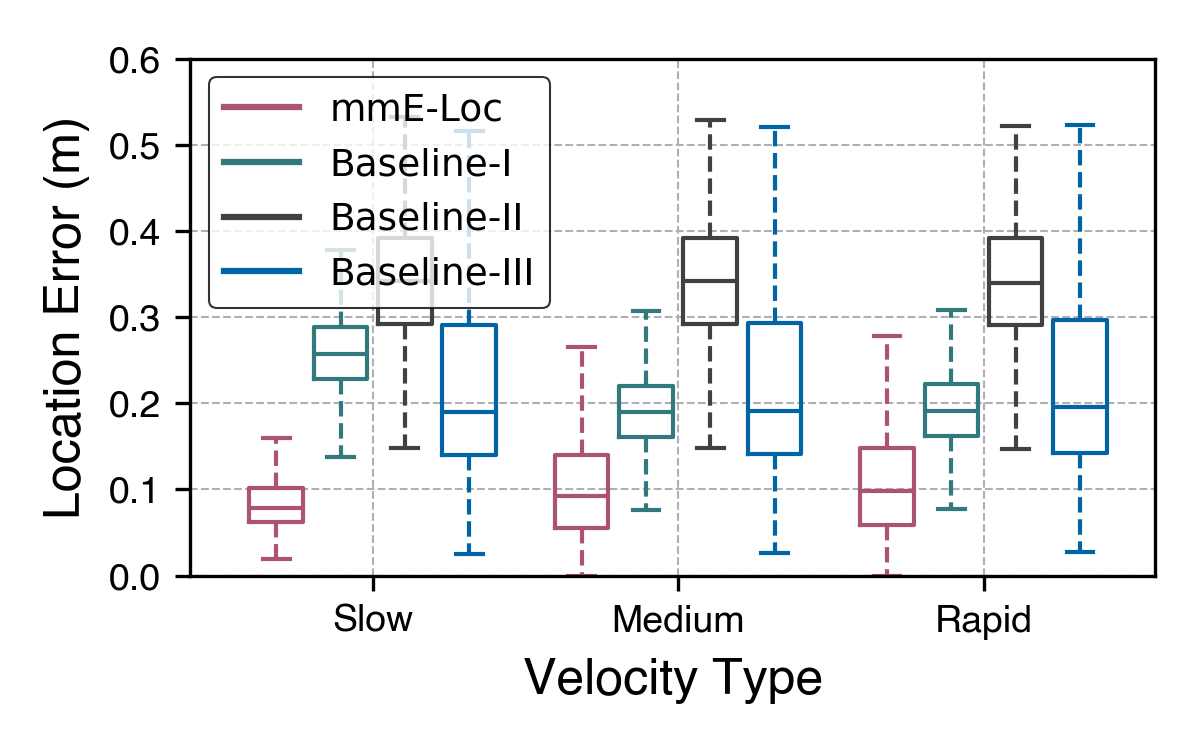}
    \caption{Impact of Velocity}
    \label{fig:velocity}
  \end{minipage}
  \hfill
  \vspace{-0.7cm}
\end{figure*}

\vspace{-0.3cm}
\textbf{End-to-end latency.}
We evaluate end-to-end latency, including the \textit{CCT} and \textit{GAJO} phases.
As shown in \fig \ref{fig:Overallperf}c, mmE-Loc achieves 5.15$ms$ latency indoors, outperforming the four baselines at 10.19$ms$, 15.12$ms$, 15.52$ms$, and 31.2 $ms$, respectively. 
In outdoor scenarios with higher complexity, baseline latencies increase, and mmE-Loc maintains the lowest latency, outperforming them by 62.5\%, 69.6\%, 72.7\%, and 83.3\%. 
Baselines face increased latency due to environmental noise and more optimization parameters.
The mmE-Loc tightly couples the event camera and mmWave radar, introducing the \textit{CCT} module to leverage temporal consistency and periodic micro motion for drone detection.
The \textit{GAJO} module and the adaptive optimization method are then used to exploit spatial complementarity, further reducing latency.


\vspace{-0.6cm}
\subsection{System Robustness Evaluation} \label{5.3}
To demonstrate the versatility and robustness of mmE-Loc, we conduct experiments under various conditions.

\textbf{Impact of Drone Type.} 
We evaluate the impact of different drone types under controlled indoor conditions, using a DJI Mini 3 Pro (\fig \ref{setup}c) and a DJI Mavic 2 (\fig \ref{setup}d), the results are presented in \fig \ref{fig:drone_type}. 
The results of average errors show that mmE-Loc outperforms all baselines with the DJI Mini 3 Pro.
With the DJI Mavic 2, Baseline-II and -III show larger localization errors due to changes in drone geometry affecting depth estimation. 
mmE-Loc's localization error of 0.135$m$ remains within an acceptable range, which outperforms all baselines with 0.222$m$, 0.639$m$, and 0.403$m$, demonstrating its effectiveness across different drone types.
\textbf{Impact of Environment \& Illumination.}
The performance of mmE-Loc in different environments with varying lighting conditions (\fig \ref{setup}e and \fig \ref{setup}f) is shown in \fig \ref{fig:illumination}. 
Although event cameras have a high dynamic range, drone-generated events are still affected by illumination. 
Baseline-II and -III experience increased errors as illumination decreases due to deteriorating depth estimation. 
In comparison, mmE-Loc sustains a relatively low average error even under low illumination, recording an average error of 0.103$m$ and a maximum error of 0.27$m$.
mmE-Loc's consistent performance across different lighting conditions, without requiring pre-training or prior knowledge, makes it a versatile solution.


\textbf{Impact of Background Motion.}
We test mmE-Loc with dynamic background motion, as shown in \fig \ref{fig:background}. 
The intensity of dynamic background motion is controlled by deploying other moving objects in the scene (\eg, balls).
All baselines show higher localization errors with increasing background motion due to misidentification of the radar and camera.
Despite this, mmE-Loc maintains an average error of 0.129$m$ in the most challenging scenarios.
This is achieved through mmE-Loc’s consistency-instructed measurement filter, which uses the drone’s periodic micro-motions of propeller rotation to distinguish it from other objects.
\textbf{Impact of Distance.}
We investigate the impact of drone-to-platform distance indoors with a DJI Mini 3 Pro (0.25$m$ $\times$ 0.36$m$ $\times$ 0.07$m$). 
\fig \ref{fig:distance} shows the results, categorized into Near (< 3$m$), Normal (3$m$ $\sim$ 6$m$), and Far (> 6$m$) distances. 
As distance increases, all methods show higher errors. 
mmE-Loc achieves an average error of 0.102$m$ for far distances, outperforming other baselines. 
mmE-Loc leverages mmWave radar for depth information and an event camera for high-resolution 2D imaging, effectively overcoming the low spatial resolution of mmWave radar and the scale uncertainty of the event camera.
Acknowledging the importance of precise landings, mmE-Loc complements existing solutions.
Working alongside RTK and visual markers, it enhances the reliability and accuracy of localization in real-world scenarios.

\begin{figure*}
\setlength{\abovecaptionskip}{-0.1cm} 
\setlength{\belowcaptionskip}{-0.25cm}
\setlength{\subfigcapskip}{-0.2cm}
    \centering
        \subfigure[Effectiveness of Sensor Fusion]{
            \centering
            \includegraphics[width=0.5\columnwidth]{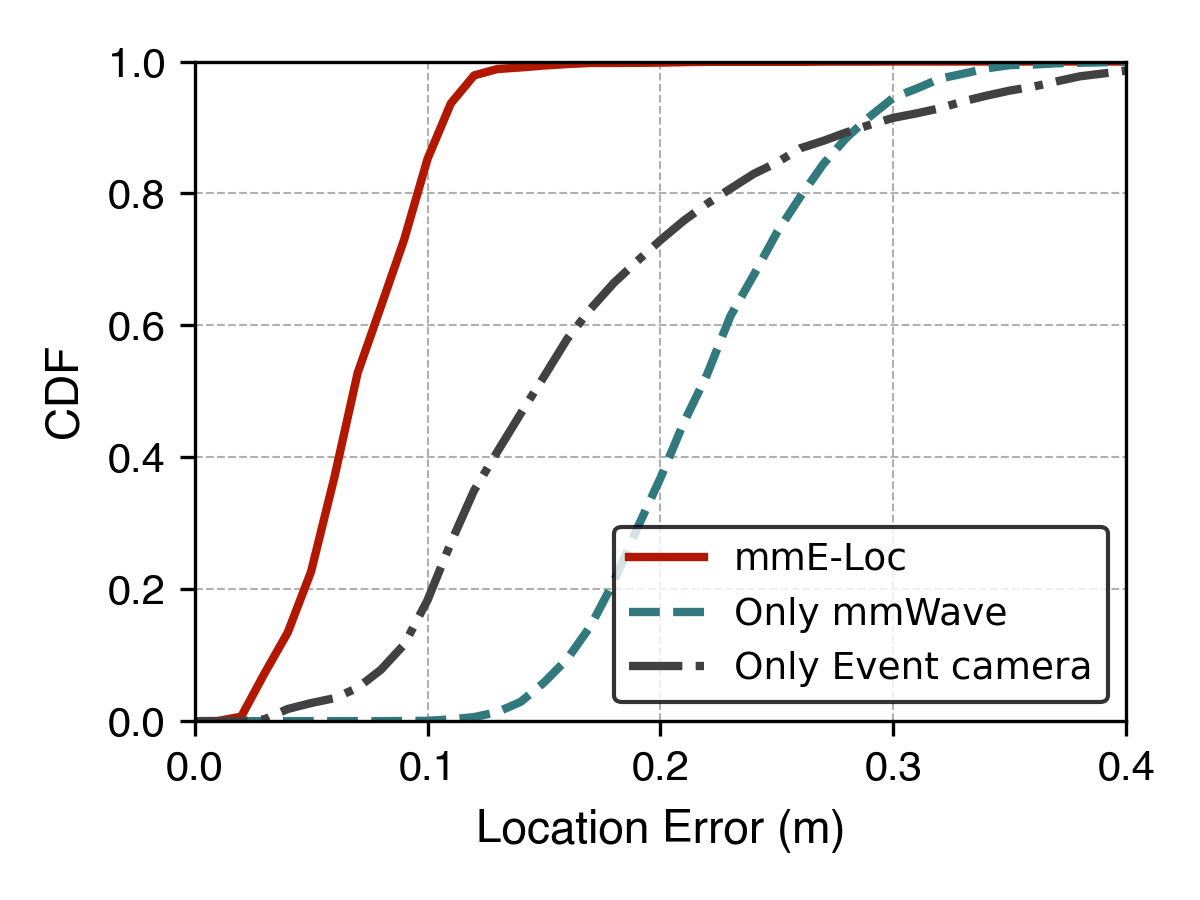}
            \label{fig:fusion}
        }%
        \subfigure[Impact of Different Module]{
            \centering
            \includegraphics[width=0.5\columnwidth]{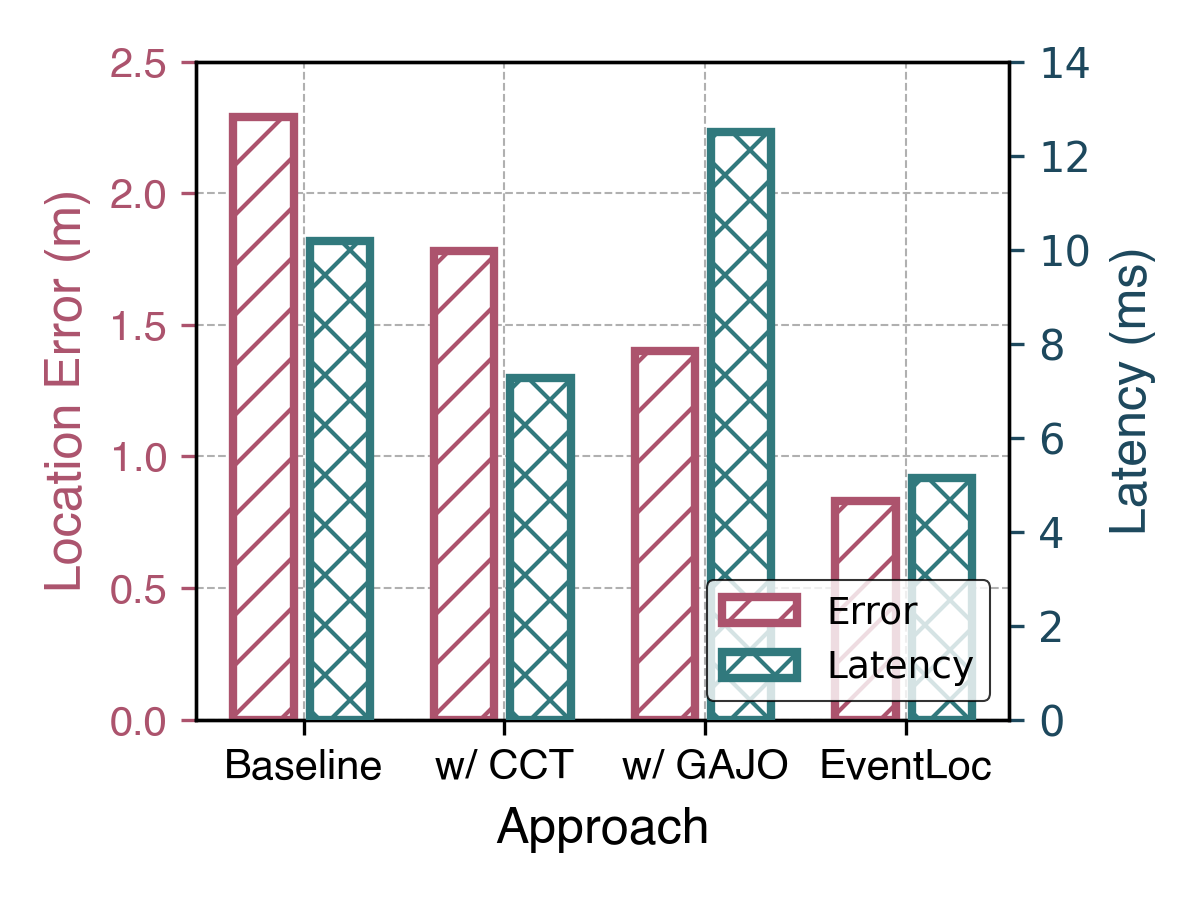}
            \label{fig:module}
        }%
        \subfigure[Performance of \textit{CCT}]{
            \centering
            \includegraphics[width=0.5\columnwidth]{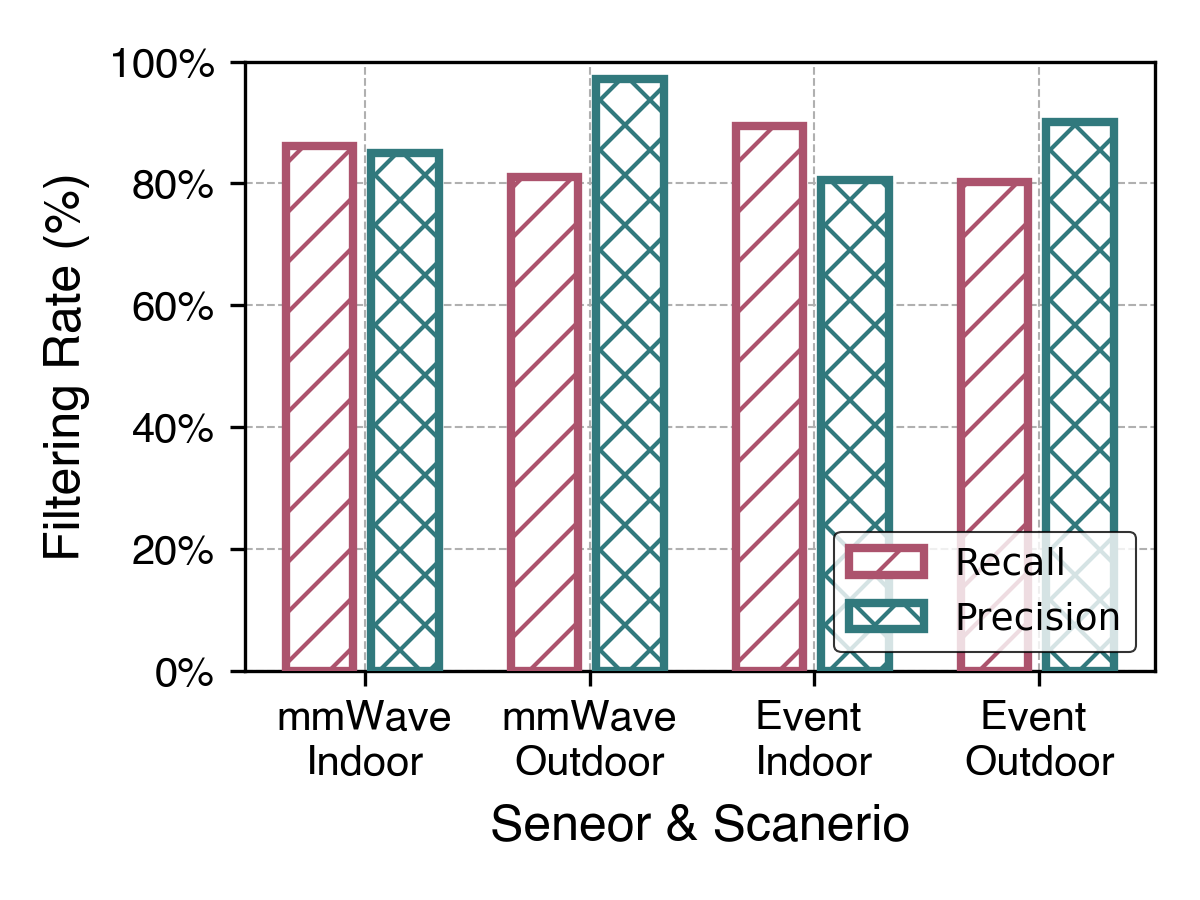}
            \label{fig:CCT}
        }%
        \subfigure[Performance of \textit{GAJO}]{
            \centering
            \includegraphics[width=0.5\columnwidth]{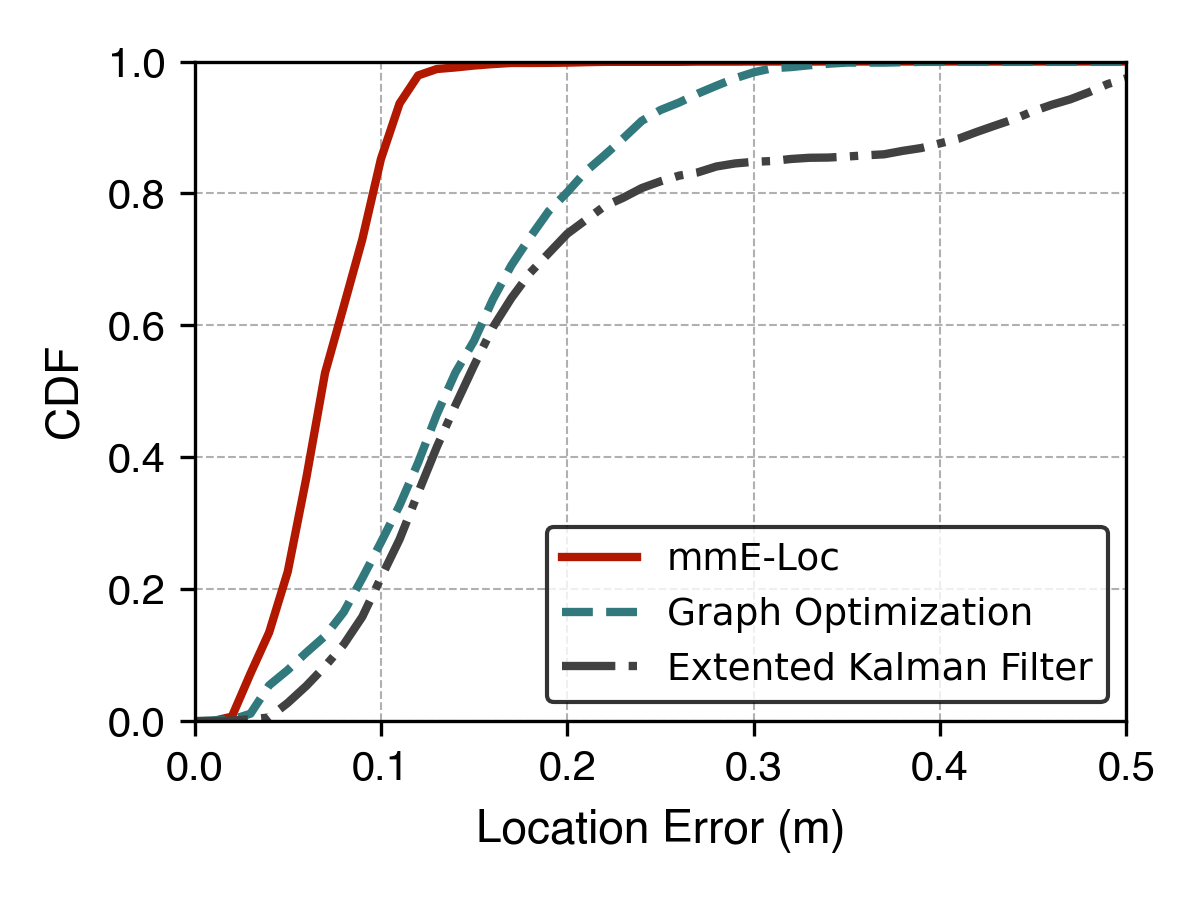}
            \label{fig:GAJO}
        }%
    \caption{Ablation Study.}
    \label{fig:ablation}
    \vspace{-0.3cm}
\end{figure*}

\begin{figure*}
\setlength{\abovecaptionskip}{-0.2cm} 
\setlength{\belowcaptionskip}{-0.15cm}
\setlength{\subfigcapskip}{-1cm}
  \begin{minipage}[t]{0.695\columnwidth}
    \centering
    \includegraphics[width=1\columnwidth]{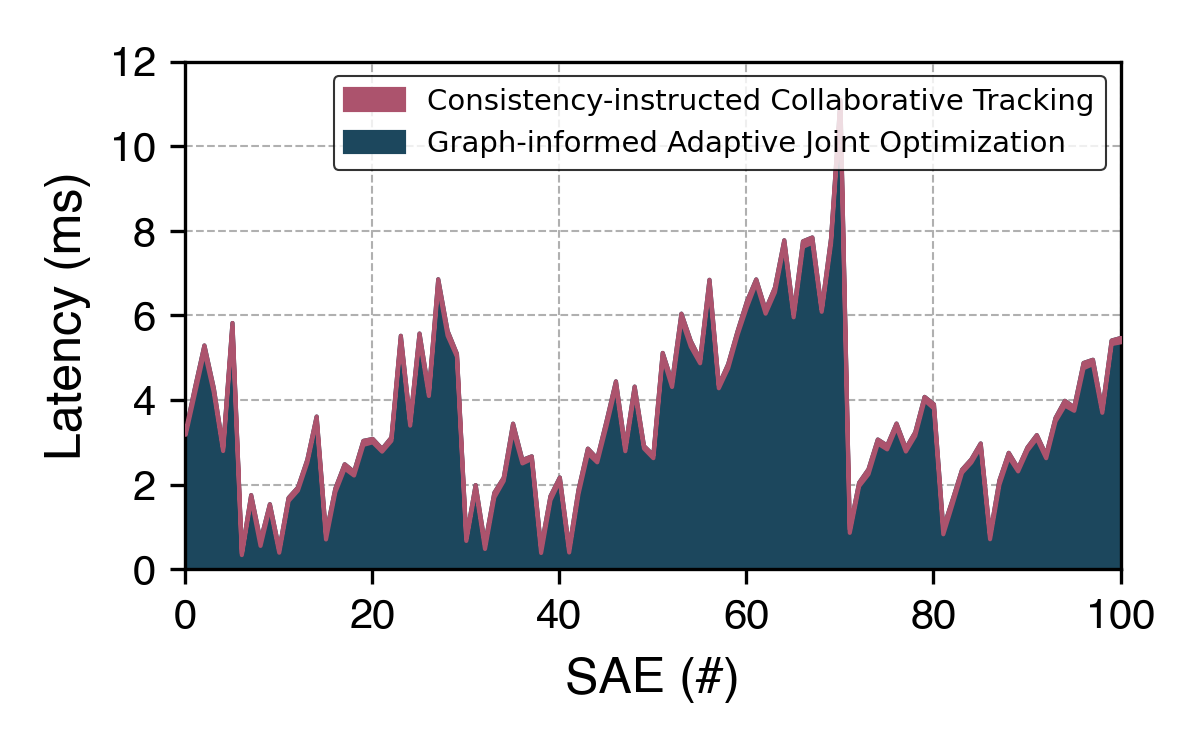}
    \caption{System Latency}
    \label{fig:latency}
  \end{minipage}
  \begin{minipage}[t]{0.692\columnwidth}
    \centering
    \includegraphics[width=1\columnwidth]{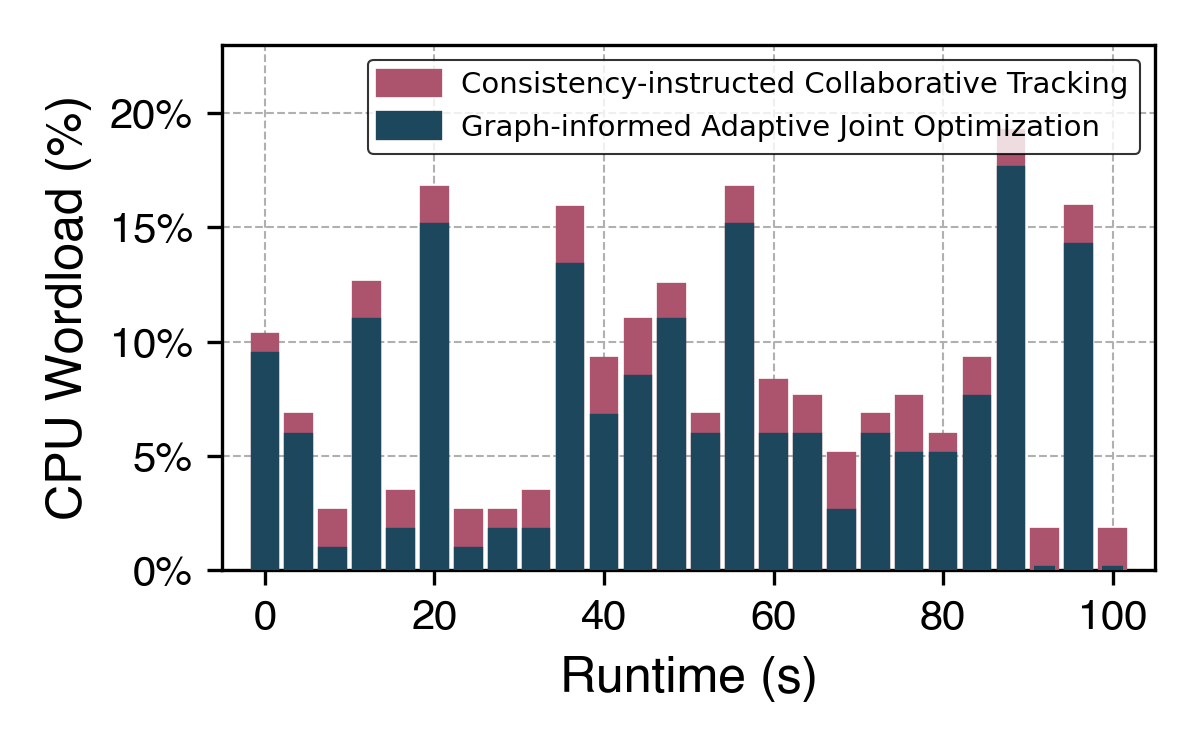}
    \caption{CPU Workload}
    \label{fig:cpu}
  \end{minipage}
  \begin{minipage}[t]{0.692\columnwidth}
    \centering
    \includegraphics[width=1\columnwidth]{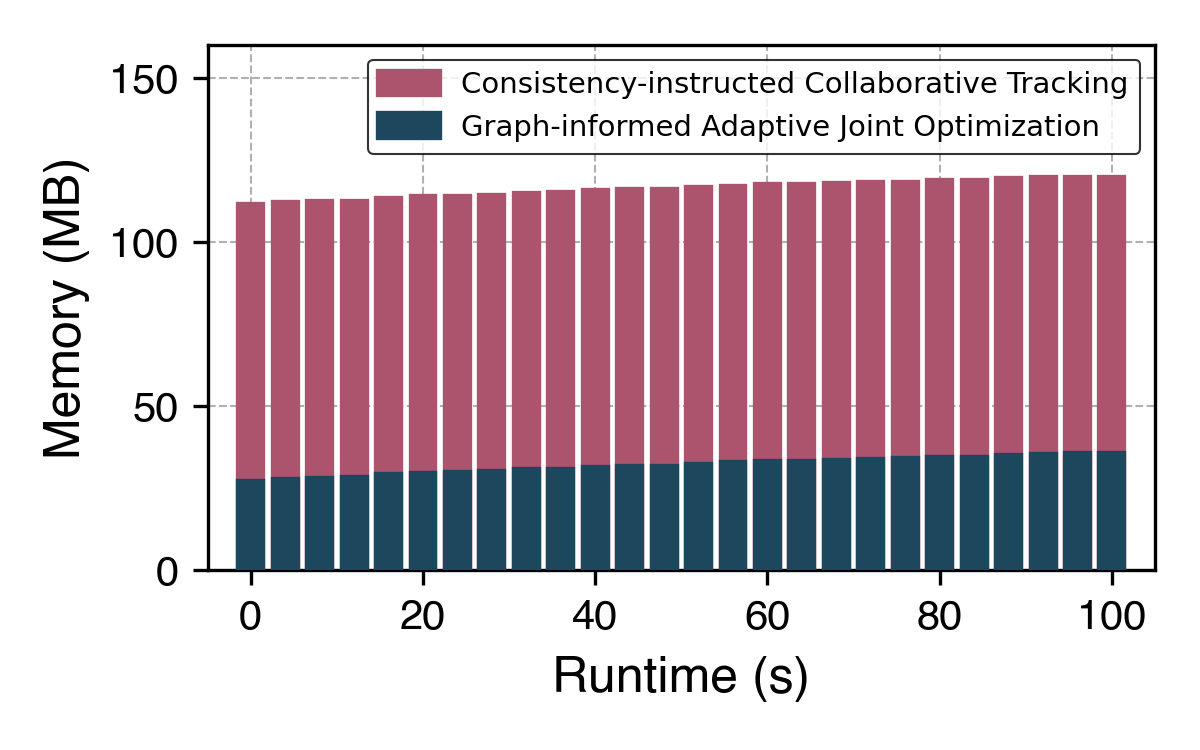}
    \caption{Memory Usage}
    \label{fig:memory}
  \end{minipage}
  \hfill
  \vspace{-0.8cm}
\end{figure*}
\textbf{Impact of Occlusion.}
We validate mmE-Loc's robustness with partially occluded drones. 
The occlusion is controlled by partially obstructing the drone within FoV of the event camera.
\fig \ref{fig:occlusion} shows that with 25\% occlusions, mmE-Loc maintains high performance with an average error of 0.094$m$. 
At 50\% occlusion, the average error of mmE-Loc increases to 0.12$m$. 
Baseline-II and -III show larger errors due to incorrect depth estimation caused by occlusion. 
mmE-Loc harnesses the strengths of both modalities in accurately tracking the drone’s location, even under partial occlusion.

\textbf{Impact of Drone Velocity.}
We further evaluate mmE-Loc's robustness under different drone velocities in \fig \ref{fig:velocity}. 
Velocities are categorized as slow ($v < 0.5m/s$), medium ($0.5m/s \leq v < 1m/s$), and rapid ($1m/s \leq v < 1.5m/s$), corresponding to various drone landing stages. 
Although error of all methods increased with speed, mmE-Loc maintained an average error of 0.11$m$ even in rapid speed scenarios, outperforming baselines by 44\%, 68.5\%, and 52.4\%.

\vspace{-0.1cm}
\subsection{Ablation Study} \label{5.4}
We experimentally analyze core components of mmE-Loc, focusing on enhancements contributed by each component to overall system.

\textbf{Effectiveness of Multi-modal Fusion.}
We demonstrate the superiority of fusing radar and event cameras over using each sensor individually. 
\fig \ref{fig:ablation}a shows that the fusion-based approach significantly outperforms both radar-only and event camera-only methods in terms of location error. 
mmE-Loc outperforms event-only approach by 63.6\% and exceeds radar-only method by 52.9\%.

\textbf{Contributions of each module.}
We investigate the contributions of \textit{CCT} and \textit{GAJO} to mmE-Loc by gradually integrating them with the event camera into the baseline system (\ie, the radar-only-based method) and assessing localization accuracy and end-to-end latency. 
\fig \ref{fig:ablation}b illustrates that without these modules, the baseline method achieves a localization error of 0.229$m$ and latency of 10.19$ms$.
Integrating the event camera with the \textit{CCT} module reduces the localization error to 0.178$m$ and decreases latency to 7.27$ms$. 
Integrating the event camera with \textit{GAJO} further reduces the error to 0.139$m$, although the delay increases due to the absence of an efficient detection mechanism. 
Finally, integrating both \textit{CCT} and \textit{GAJO} minimizes both the error and latency.

\textbf{Performance of \textit{CCT}.}
We tested \textit{CCT}'s filtering performance on mmWave and event data in both indoor and outdoor scenarios. 
In \fig \ref{fig:ablation}c, higher recall signifies more drone-triggered events preserved, while higher precision indicates more background events removed. 
In indoor conditions, \textit{CCT} achieves recalls over 86\% for mmWave and 89\% for event, with precision above 85\% and 80\%, respectively. 
In outdoor conditions, recall and precision remain above 80\% for both types, demonstrating \textit{CCT}'s effectiveness.

\textbf{Performance of \textit{GAJO}.}
We also compare the performance of different multi-modal fusion strategies. 
Specifically, we evaluate mmE-Loc against two widely used approaches: the extended Kalman filter (EKF) and Graph Optimization (GO). 
As shown in \fig \ref{fig:ablation}d, mmE-Loc enhances localization performance by over 57.9\% compared to EKF and 47.1\% compared to GO, due to its tightly coupled multi-modal fusion and adaptive optimization method.
\vspace{-0.3cm}
\subsection{System Efficiency Study} \label{5.5}
\revise{

The mmE-Loc distinguishes itself from existing models and learning-based methods due to its low latency and minimal resource overhead. 
\fig \ref{fig:latency} illustrates the end-to-end latency (including delays from \textit{CCT} and \textit{GAJO} modules) throughout the localization process. 
The average end-to-end latency of mmE-Loc is around 3.58$ms$, with the \textit{CCT} module contributing an average delay of 0.09$ms$ and the \textit{GAJO} module adding 3.49$ms$. 
During drone localization, mmE-Loc's latency may fluctuate due to the adaptive optimization method, which optimizes different sets of locations at different times. 
Nonetheless, mmE-Loc's latency remains suitable for use in flight control loops. 
\fig \ref{fig:cpu} and \fig \ref{fig:memory} indicate that CPU usage doesn't exceed 20\%, with memory usage under 120$MB$.
Meanwhile, memory usage increases as the location set size grows.
}

\begin{figure*}[t]
    \setlength{\abovecaptionskip}{0.2cm} 
    \setlength{\belowcaptionskip}{-0.5cm}
    \setlength{\subfigcapskip}{-1cm}
    \centering
        \includegraphics[width=1.95\columnwidth]{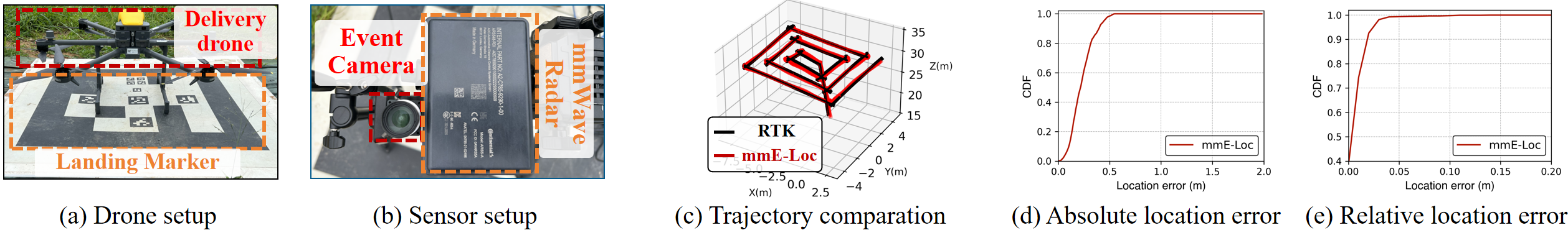}
        \vspace{-0.15cm}
    \caption{Case Study: Delivery Drone Airport. }
    \label{airport}
    \vspace{-0.15cm}
\end{figure*}


\vspace{-0.2cm}
\section{Case Study: Drone Airport}
As shown in \fig \ref{airport}a, to verify the system's usability, we conduct an experiment using a custom drone equipped with six propellers and managed by a PX4 flight controller. 
This drone is developed by a world-class delivery company exploring the feasibility of instant deliveries.
The experiment takes place at a real-world delivery drone airport.
To enable drone localization over a larger area, we employ an ARS548 mmWave radar, as depicted in \fig \ref{airport}b. 
\fig \ref{airport}c, \fig \ref{airport}d, and \fig \ref{airport}e illustrate the localization results as the drone follows a square spiral trajectory at an altitude of 30 $m$.
The results show that in real-world scenarios, mmE-Loc achieves high localization precision, maintaining a maximum absolute location error error below 0.5$m$ and relative location error error below 0.1$m$, and producing smooth trajectories that closely align with those of RTK. 
mmE-Loc has significant potential as a complementary system to RTK, aiding drone in challenging environments (\eg, urban canyons, where RTK may be compromised by signal blockage).
\vspace{-0.3cm}
\section{Related work}

\revise{
\textbf{Drone ground localization.}
Several types of work have been proposed to assist drones in localization.
$(i)$ \textit{Satellite-based systems}.
The Global Positioning System (GPS) provides $m$-level accuracy outdoors \cite{wang2023global, dong2023gpsmirror}, while Real-Time Kinematic (RTK) achieves $cm$-level precision but is costly. 
However, these satellite-based systems struggle in urban canyons \cite{wang2022micnest, xu2020edge, wang2024transformloc, wang2022h}. 
$(ii)$ \textit{Optics-based systems}.
These systems, such as motion capture, offer $cm$-level accuracy indoors but require precise calibration, making them impractical for outdoor use \cite{xue2022m4esh}. 
$(iii)$ \textit{Sensor-based systems}.
To address these issues, various sensor-based techniques are proposed, including camera \cite{he2022automatch, ben2022edge, tasneem2020adaptive}, radar \cite{sie2023batmobility, iizuka2023millisign}, LiDAR \cite{cui2024vilam, jian2024lvcp} and acoustic \cite{wang2022micnest} often combined with SLAM (Simultaneous Localization and Mapping) \cite{chen2020h, chen2017design, chen2023adaptslam, chen2015drunkwalk} or deep learning algorithms \cite{zhao2024foes, zhao2023smoothlander, hu2024seesys, zhao2024understanding, yimiao2023bifrost}, aim to improve drone localization. 
However, limited spatio-temporal resolution in these sensors affects precise, low-latency landing drone localization. 
For example, cameras and LiDARs, with low frame rates (< 50$Hz$), may miss rapid movements during frames, reducing localization accuracy \cite{jian2023path}.
Acoustic signal-based systems are highly susceptible to environmental noise, with even nearby humans impacting their stability.
Visual marker-based systems work in conjunction with downward-facing cameras mounted on drones. 
However, while these systems assist drones in obtaining their location, they do not help ground platforms track drones. 
They also are sensitive to lighting conditions due to limited dynamic range of frame cameras.

Compared to previous methods, mmE-Loc leverages a novel sensor configuration combining event camera with mmWave radar, harmonizing ultra-high sampling frequencies, to achieve superior drone ground localization accuracy with low latency.
Meanwhile, mmE-Loc is resistant to lighting variations due to its sensor configuration, with event cameras offering a high dynamic range. 
It is important to note that mmE-Loc complements, rather than replaces, existing localization solutions.
To ensure precise landing, mmE-Loc will work in conjunction with RTK and visual markers, providing a more reliable and accurate localization service.
}

\textbf{mmWave for localization and tracking.}
Millimeter-wave is highly sensitive and more accurate due to its mm-level wavelength.
mmWave radar offers high sensitivity and precision due to its sub-millimeter wavelength \cite{Harlow_2024,s23218901,lu2020smokerobustindoormapping}. 
Several mmWave radar-based solutions for drone ground localization combine signal intensity methods, but face challenges in accurately tracking the drone's center \cite{qian20203d, asi6040068, 10.1145/3678549}. 
This difficulty arises from the drone's large size (\eg, 80 $cm$ across), causing it to appear as a non-uniform blob in radar returns. 
Additionally, these methods often produce unstable results with frequent outliers, as multipath scattering can obscure the main signal and low spatial resolution of radar \cite{Li2024AHH}.
Other approaches leverage deep learning-based methods but require extensive pre-modeling and neural network training for each drone model \cite{lu2020see, zheng2023neuroradar, ALLQUBAYDHI2024100614}. 
These methods tend to struggle with tracking different drone models and perform poorly in environments not represented in training dataset. 
Several solutions integrate visual sensors to assist radar \cite{shuai2021millieye,chadwick2019distant, cho2014multi}. 
However, these approaches introduce latency due to exposure times and image processing delays.

To overcome the accuracy and latency bottlenecks, we upgrade frame cameras with event cameras to pair with mmWave radar, boosting system performance. 
By exploiting temporal consistency and drone's periodic micro-motion with \textit{CCT} module, mmE-Loc achieves impressive tracking accuracy without the need for prior knowledge, (\eg, training data or 3D models). 
The integration of the \textit{GAJO} module employing spatial complementarity and an adaptive optimization method further enables accurate drone ground localization and reduces latency to the $ms$-level.

\revise{
\textbf{Sensor Fusion Techniques.}
Sensor fusion techniques are widely used in localization \cite{li2022motion, li2023riscan, wang2023meta, xue2023towards}. 
\textit{(i) Traditional pipeline fusion methods.}  
These approaches typically integrate dense point clouds from sensors (\eg, LiDAR) to provide depth information, alongside pixel data from frame cameras \cite{he2023vi, shi2024soar}. 
However, these methods are generally limited to fusing two types of dense measurements and cannot directly accommodate the sparse data output from event cameras and mmWave radar. 
\textit{(ii) Deep learning-based fusion methods.}  
Recent advancements have explored learning-based methods for fusing mmWave radar, frame cameras, and IMUs for localization \cite{xu2021followupar, safa2023fusing}. 
While these methods offer promising results, they often require extensive labeled training data and experience performance degradation in dynamic environments \cite{ shuai2021millieye, lu2020see, ding2023push, li2023egocentric, muller2023aircraft}. 
In contrast, mmE-Loc adopts a tightly coupled fusion framework, leveraging factor graphs and adaptive optimization to jointly optimize radar and event tracking models. 
This approach provides a clearer probabilistic interpretation compared to deep learning-based methods.

}

\vspace{-0.5cm}
\section{Discussion}\label{8}
\revise{

\noindent \textbf{How does mmE-Loc relate to visual markers?}
Currently, delivery drones utilize visual markers and onboard cameras for self-localization. 
In contrast, mmE-Loc focuses on ground-based drone localization, enabling the landing pad to determine the spatial relationship between the drone and itself for precise adjustments. 
In practice, mmE-Loc operates alongside visual markers to enhance reliability and accuracy of localization service.

\noindent \textbf{Is it feasible to design an onboard sensor system for drone landing localization?}
It's theoretically feasible. 
However, designing an onboard system for drone localization using an event camera and mmWave radar presents several challenges. 
Given the continuous motion of the drone, the system must address:  
$(i)$ motion compensation for event data,  
$(ii)$ reliable feature extraction and matching within the event stream despite its lack of semantic information, and  
$(iii)$ the sparsity of radar measurements and noise induced by specular reflections, diffraction, and multi-path effects.


\noindent \textbf{How to manage simultaneous drone landings?}
As multiple drones land simultaneously, mmE-Loc will initialize multiple trackers within the event tracking model to track each drone, associate the results with radar tracking model, and subsequently perform localization and optimization for each drone individually.

\noindent \textbf{How does network latency impact the system, and how can potential delay-related issues be addressed?}
mmE-Loc is designed to assist ground platforms in locating drones and guiding them to land accurately at designated spots.
Network latency may impact the location update rate for the drone. 
To mitigate this issue, airports should deploy access points near the landing pad and utilize multiple wireless links, including Wi-Fi and cellular networks, to ensure reliable and fast communication.
Additionally, an optics-based communication system could be integrated into the system to ensure a high communication frequency \cite{wang2024towards}.

\noindent \textbf{How does strong light affect system performance?}
Lighting conditions primarily affect event cameras.
Strong light has minimal impact on event cameras, as they detect only changes in light intensity. 
In contrast, weak illumination affects performance more significantly since intensity changes are less pronounced.
Our experimental results on illumination effects (Fig. 12) confirm that weak illumination introduces greater errors than strong light.

}

\vspace{-0.2cm}
\section{Conclusion}\label{7}

This paper explores a novel sensor configuration combining event camera and mmWave radar, harmonizing ultra-high sampling frequencies, and proposes mmE-Loc, a ground localization system for drone landings, achieving $cm$  accuracy and $ms$ latency. 
The innovation of mmE-Loc lies in two aspects: 
$(i)$ a Consistency-Instructed Collaborative Tracking module that leverages cross-modal \textit{temporal consistency} for accurate drone detection, and $(ii)$ a Graph-Informed Adaptive Joint Optimization module that boosts localization performance and reduces latency by utilizing cross-modal \textit{spatial complementarity}. 
Extensive evaluations conducted across various scenarios demonstrate performance of mmE-Loc.

\section{Acknowledgement}
We sincerely thank the anonymous shepherd for constructive comments and feedback in improving this work. 
This paper was supported by Yunnan Forestry and Grassland Science and Technology Innovation Joint Special Project (grant NO. 202404CB090017), Natural Science Foundation of China under Grant 62371269, Guangdong Innovative and Entrepreneurial Research Team Program (2021ZT09 L197), Meituan Academy of Robotics Shenzhen.

\newpage
\balance

\bibliographystyle{unsrt}
\bibliography{reference}










\end{document}